\documentclass[letterpaper]{article} 
\usepackage{aaai25}
\usepackage{times}  
\usepackage{helvet}  
\usepackage{courier}  
\usepackage[hyphens]{url}  
\usepackage{graphicx} 
\usepackage{natbib}  
\usepackage{caption} 
\usepackage{algorithm}
\usepackage{algorithmic}
\usepackage{newfloat}
\usepackage{listings}
\DeclareCaptionStyle{ruled}{labelfont=normalfont,labelsep=colon,strut=off} 
\floatstyle{ruled}
\newfloat{listing}{tb}{lst}{}
\floatname{listing}{Listing}
\usepackage{subcaption}
\usepackage{amsmath, amssymb, amsthm,mathrsfs}
\usepackage{csquotes}
\usepackage{booktabs}
\usepackage{multirow}
\usepackage[dvipsnames]{xcolor} 
\DeclareMathOperator{\Tr}{Tr}

\newtheorem{defn}{Definition}

\usepackage{fix-cm}
\usepackage[colorlinks=true, citecolor=Aquamarine,linkcolor=orange, urlcolor=blue, pdfborder={0 0 0}]{hyperref}

\title{THESAURUS: Contrastive Graph Clustering by Swapping Fused Gromov-Wasserstein Couplings}
\author {
	Bowen Deng\textsuperscript{\rm 1,\rm 2}\equalcontrib,
	Tong Wang\textsuperscript{\rm 1}\equalcontrib,
	Lele Fu\textsuperscript{\rm 1,\rm 2},
	Sheng Huang\textsuperscript{\rm 1,\rm 2},
	Chuan Chen\textsuperscript{\rm 1}\thanks{Corresponding authors.},
	Tao Zhang\textsuperscript{\rm 2}\footnotemark[2],
}
\affiliations {
	\textsuperscript{\rm 1}School of Computer Science and Engineering, Sun Yat-sen University, Guangzhou, China\\
	\textsuperscript{\rm 2}School of Systems Science and Engineering, Sun Yat-sen University, Guangzhou, China\\
	\{dengbw3, wangt, full, huangs\}@mail2.sysu.com, \{chenchuan,zhangt358\}@mail.sysu.edu.cn
}

\begin{document}

\maketitle

\begin{abstract}
	Graph node clustering is a fundamental unsupervised task. Existing methods typically train an encoder through self-supervised learning and then apply K-means to the encoder output. Some methods use this clustering result directly as the final assignment, while others initialize centroids based on this initial clustering and then finetune both the encoder and these learnable centroids. However, due to their reliance on K-means, these methods inherit its drawbacks when the cluster separability of encoder output is low, facing challenges from the Uniform Effect and Cluster Assimilation. We summarize three reasons for the low cluster separability in existing methods: \textbf{(1)} lack of contextual information prevents discrimination between similar nodes from different clusters; \textbf{(2)} training tasks are not sufficiently aligned with the downstream clustering task; \textbf{(3)} the cluster information in the graph structure is not appropriately exploited. To address these issues, we propose conTrastive grapH clustEring by SwApping fUsed gRomov-wasserstein coUplingS (THESAURUS). Our method introduces semantic prototypes to provide contextual information, and employs a cross-view assignment prediction pretext task that aligns well with the downstream clustering task. Additionally, it utilizes Gromov-Wasserstein Optimal Transport (GW-OT) along with the proposed prototype graph to thoroughly exploit cluster information in the graph structure.
	To adapt to diverse real-world data, THESAURUS updates the prototype graph and the prototype marginal distribution in OT by using momentum.
	Extensive experiments demonstrate that THESAURUS achieves higher cluster separability than the prior art, effectively mitigating the Uniform Effect and Cluster Assimilation issues.
\end{abstract}

%
 \begin{links}
	    \link{Code}{https://github.com/bwdeng20/THESAURUS}
	 \end{links}

\section{Introduction}

Graph node clustering
\citep{wang2024OverviewAdvancedDeep,liu2023SurveyDeepGraph} is a fundamental
unsupervised task. Recently, methods based on Graph Self-Supervised Learning (Graph SSL) \citep{liu2022GraphSelfSupervisedLearning}
have become predominant \citep{liu2023SurveyDeepGraph}. Despite their
success, 
these methods, e.g., Dink-Net \citep{liu2023DinkNetNeuralClustering}, heavily rely on K-means \citep{lloyd1957LeastSquaresQuantization,macqueen1967MethodsClassificationAnalysis} to guide
the representation learning process and/or to get the final clustering results, and thus inherit the shortcomings of K-means. Clusters that contain significantly more samples than others are called majority clusters, and conversely, those with fewer samples minority clusters.
When the input node representations exhibit low cluster separability, K-means results may show \textbf{(1) Uniform Effect}: samples from majority clusters being assigned to neighboring minority clusters \citep{xiong2009KMeansClusteringValidation},
and \textbf{(2) Cluster Assimilation}: minority clusters being merged into neighboring majority clusters \citep{lu2021SelfAdaptiveMultiprototypeBasedCompetitive}. In contrast, high cluster separability, an ideal clustering outcome characterized
by large inter-cluster and small intra-cluster distances,
can alleviate these two issues \citep{lu2021SelfAdaptiveMultiprototypeBasedCompetitive}, as demonstrated by the Dink-Net finetune effect experiment presented below.

\subsection{Uniform Effect \& Cluster Assimilation in Dink-Net}
\begin{figure}[t]
	\centering
	\begin{subfigure}{0.95\columnwidth}
		\includegraphics[width=1\linewidth]{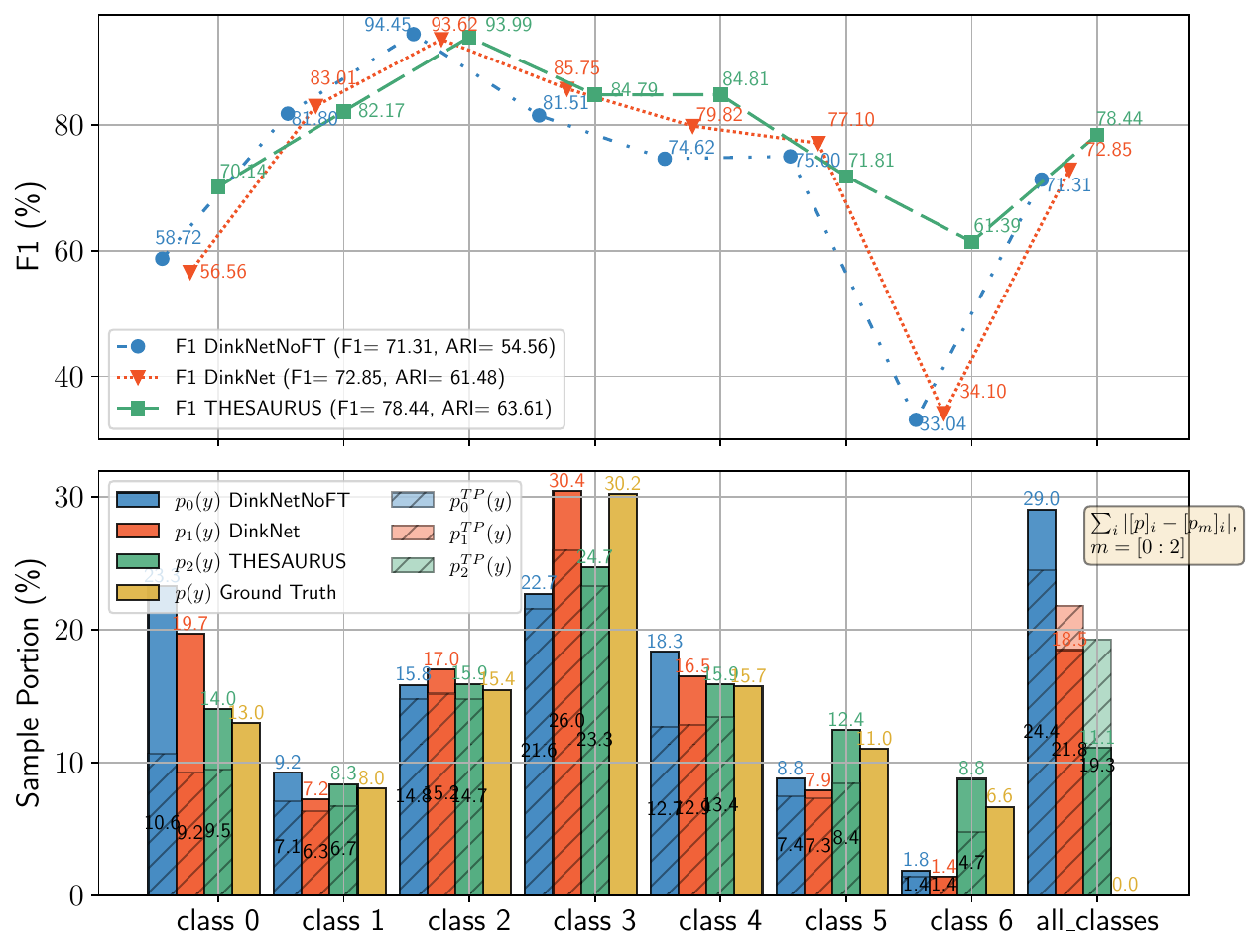}
		\caption{The F1 and cluster label histograms}
		\label{fig:DinkNetFintuneEffect-F1} 
	\end{subfigure}
	\begin{subfigure}{0.95\columnwidth}
		\includegraphics[width=1\linewidth]{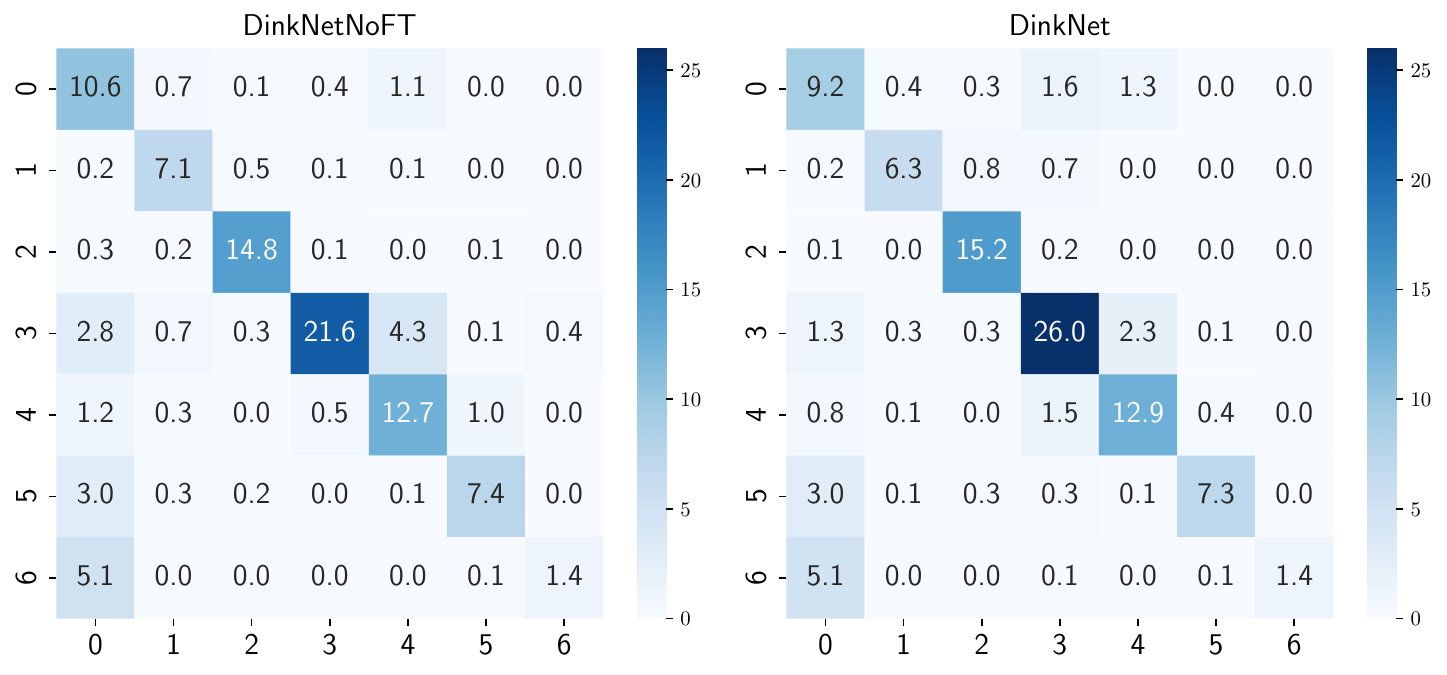}
		\caption{The confusion matrices before/after finetune}
		\label{fig:DinkNetFintuneEffect-CM}
	\end{subfigure}
	\caption{The effect of separability-oriented finetune of Dink-Net on the Cora dataset. \textbf{(a)} The \textbf{top row} illustrates the F1 scores for each class before and after finetune, as well as the average F1 score over all classes. The
		\textbf{second row  }shows the distribution of predicted labels from three models, along with the ground-truth labels. It also presents the distribution of predicted labels for true-positive (TP) samples, denoted as \(p_{i}^{TP}(y),i \in \{0,1,2\}\). The final set of bars shows the differences between the predicted and ground-truth distributions. \textbf{(b)} displays the confusion matrices (\%) of Dink-Net before and after finetune, normalized by the number of nodes.}
	\label{fig:DinkNetFintuneEffect}
\end{figure}

The current state-of-the-art (SOTA) model, Dink-Net, is pretrained by distinguishing the original data and the randomly corrupted and shuffled data. We denote the pretrained Dink-Net as Dink-Net-NoFT. After pretraining, K-means is employed to cluster the Dink-Net-NoFT output, initializing the centroids $\left\{ \mathbf{c}_{i}\right\} _{i=0}^{C-1}$. In the later finetune stage, the encoder and  centroids are adjusted to enhance cluster separability by minimizing the dilation loss  $\mathcal{L}_{d}=\frac{-1}{(C-1)C}\sum_{i}\sum_{i\neq i}\|\mathbf{c}_{i}-\mathbf{c}_{j}\|_{2}^{2}$ and shrink loss $\mathcal{L}_{s}=\frac{1}{BC}\sum_{i=1}^{B}\sum_{j=0}^{C-1}\|\mathbf{z}_{i}-\mathbf{c}_{j}\|_{2}^{2}$, 
where $B$ is the node batch size and $\mathbf{z}_{i}$ is the representation. For a node $i$, its predicted cluster
is $\hat{\mathbf{y}}_{i}=\arg\min_{j}\|\mathbf{z}_{i}-\mathbf{c}_{j}\|_{2}$.
Fig. \ref{fig:DinkNetFintuneEffect} shows the finetune impact on Dink-Net. 

Before finetune, two phenomena are observed on the Cora dataset. \textbf{(1) Uniform Effect}: the largest cluster (cluster 3) has many external edges with the much smaller clusters 0 and 4, causing them to be close in the embedding space. This is evidenced in Fig. \ref{fig:DinkNetFintuneEffect-CM}, where many nodes from cluster 3 are misclassified into clusters 4 and 0. \textbf{(2) Cluster Assimilation}: The smallest cluster (cluster 6) has the most external edges with cluster 0, leading to its merging into cluster 0. 

After finetuning towards high separability, some of the nodes from cluster 3 but misclassified into clusters 4 and 0 are returned to cluster 3, resulting in a significant improvement in the F1 score for cluster 3. However, since the cluster separability is still insufficient, the smallest cluster (cluster 6) remains merged into cluster 0, leading to a low F1 score on class 6 and a low Macro-F1 score. This experiment underscores the critical role of high cluster separability and reveals that even the current SOTA struggles to achieve sufficient cluster separability to effectively address the challenges of Uniform Effect and Cluster Assimilation.

\subsection{Current Model Limitations and Contributions}
There are three \textbf{limitations (Ls)} impeding current methods achieving high cluster separability. \textbf{L1: Lack of contextual information hinders distinguishing synonymous nodes (i.e., similar nodes from different classes).} When a graph has many inter-class edges, low-pass (learnable) graph filters, e.g., GCN \citep{kipf2017semisupervised}, tend to generate similar embeddings for neighboring nodes from different classes \citep{chen2020a}. Distinguishing such nodes based solely on embedding distances is challenging. Just as synonyms in a thesaurus need context to be properly understood, \enquote{childish} implies immaturity in \enquote{Stop being so childish!} while \enquote{childlike} conveys innocence in \enquote{She has a childlike wonder.} Nodes also require contextual information for accurate clustering. Current methods depend only on the embedding distance and no contextual details are provided. As a result, they fail to differentiate between closely embedded nodes from different classes, e.g., mixing nodes from clusters 0, 5, and 6 in Fig. \ref{fig:DinkNetFintuneEffect}. 
\textbf{L2: Training tasks are not well aligned with downstream clustering task.} The alignment between pretext and downstream tasks is crucial for SSL \citep{lee2021PredictingWhatYou,wei2021AligningPretrainingDetection}. However, current pretext tasks often lack this alignment. \textbf{(1)} SDCN \citep{bo2020StructuralDeepClustering} reconstructs attributes and adopts a dual self-supervised strategy derived from DEC \citep{xie2016UnsupervisedDeepEmbedding}. DFCN \citep{tu2021DeepFusionClustering} reconstructs both attributes and structures with a DEC-like triplet self-supervised task. 
The reconstruction tasks do not optimize cluster separability and fail to align closely with clustering. DEC-style tasks use K-means to cluster pretrained representations initially. However, the pretrained representations often exhibit low separability, incurring Uniform Effect and Cluster Assimilation. Since finetune only refines the initial clustering, substantial improvements in addressing these issues are not available via DEC-style tasks.
\textbf{(2)} Regarding the contrastive ones, DCRN \citep{liu2022DeepGraphClustering} and HSAN \citep{liu2023HardSampleAware} only preserve self-correlations, not aligned with clustering.
SCGC \citep{liu2023SimpleContrastiveGraph} and S$^3$GC \citep{devvrit2022S3GCScalableSelfSupervised} treat neighbors as positive pairs, but neighbors are not always of the same class, introducing clustering noise. 
Although the finetune task of Dink-Net aligns with the clustering objective, the unrelated pretrain task limits the representation separability. Thus, Dink-Net finetune cannot resolve all challenges, as shown in Fig. \ref{fig:DinkNetFintuneEffect}.
\textbf{L3: The cluster information in graph structure is not appropriately extracted.} Existing methods primarily integrate structure information into embeddings via encoding with GCNs \citep{kipf2017semisupervised}. However, over-smoothing and over-squashing \citep{nguyen2023RevisitingOversmoothingOversquashing} may hurt structure information. 
HSAN  \citep{liu2023HardSampleAware}
processes structure through a linear layer, yet it lacks permutation invariance, unduly emphasizing node indices over
structure information. DFCN \cite{tu2021DeepFusionClustering} and DCRN
\citep{liu2022DeepGraphClustering} reflect the structure information through adjacency reflection loss. SCGC \citep{liu2023SimpleContrastiveGraph} and S$^3$GC \citep{devvrit2022S3GCScalableSelfSupervised} take neighbors as positive samples and maximizes their similarities. These four methods implicitly treat neighbors as belonging to the same class, misleading the clustering on adjacent nodes from different classes.

To address the above limitations, we propose a novel contrastive graph clustering method, THESAURUS. \textbf{(1)} It establishes semantic prototypes in the embedding space, each representing a semantic category. The relationships between one node and these prototypes constitute its context. \textbf{(2)} Inspired by SwAv \citep{caron2020UnsupervisedLearningVisual}, THESAURUS considers semantic prototypes as centroids and learns by predicting the node clustering assignments across different data augmentation views. \textbf{(3)} To explore the structure cluster information, we encode the relationships between prototypes as prototype graph, and then match it with the data graph using GW-OT \citep{memoli2011GromovWassersteinDistances,peyre2019ComputationalOptimalTransport}. \textbf{(4)}  To exploit structure and attribute information comprehensively, designs (2) and (3) are unified by our Task and Structure Alignment (TSA) module based on Fused
Gromov-Wasserstein OT (FGW-OT) \cite{titouan2019optimal}. \textbf{(5)} A momentum module for prototype graph and marginal distribution is developed for data adaptability.

Our main contributions are as follows. 
\textbf{(1)} We identify that prior methods has insufficient cluster separability and face the Uniform Effect and Cluster Assimilation challenges. 
\textbf{(2)} We propose a novel graph contrastive learning framework THESAURUS, which leverages semantic prototypes to provide contextual information. We design the TSA module to align the pretext task to clustering and exploit the cluster information in graph structure. We develop a momentum strategy for the prototype graph and prototype marginal distribution for data adaptability. 
\textbf{(3)} Extensive experiments demonstrate that THESAURUS achieves high cluster separability and significantly outperforms existing methods.

\section{Related Work}
\subsection{Deep Graph Clustering}
Early graph clustering methods use Autoencoder (AE) and Graph Autoencoder (GAE) \citep{kipf2016VariationalGraphAutoencoders} for feature extraction, followed by K-means or spectral clustering \citep{cao2016DeepNeuralNetworks,wang2017mgae,pan2018AdversariallyRegularizedGraph}. Later methods such as DAEGC \citep{wang2019AttributedGraphClustering}, SDCN \citep{bo2020StructuralDeepClustering}, AGCN \citep{peng2021AttentiondrivenGraphClustering}, and DFCN \citep{tu2021DeepFusionClustering} incorporate DEC-style tasks to better align with clustering objectives. With the advent of graph contrastive learning, contrastive graph clustering \citep{li2024ContrastiveDeepNonnegative} gained popularity. AGE \citep{cui2020AdaptiveGraphEncoder} uses dynamic positive and negative sample pairs constructed with pair similarities, DCRN \citep{liu2022DeepGraphClustering} introduces DICR loss to reduce cross-view correlations between nodes, SCGC \citep{liu2023SimpleContrastiveGraph} maximizes neighbor similarity, and S$^3$GC \citep{devvrit2022S3GCScalableSelfSupervised} optimizes a one-layer GNN with InfoNCE-style loss \citep{oord2019RepresentationLearningContrastive}. DinkNet \citep{liu2023DinkNetNeuralClustering} maximizes differences between original and adversarial data, akin to maximizing the JSD lower bound of mutual information \citep{shrivastava2023CLIPLiteInformationEfficient}, and then finetunes towards cluster separability via minimizing \(\mathcal{L}_d\) and \(\mathcal{L}_s\).
\subsection{Optimal Transport}
Optimal transport (OT) \citep{monge1781memoire,villani2009OptimalTransportOld} is a mathematical framework for measuring distances between distributions, finding the most cost-efficient way to transform one distribution into another. It has gained prominence in machine learning for tasks like domain adaptation \citep{courty2017JointDistributionOptimal} and generative modeling \citep{tolstikhin2018WassersteinAutoencoders}. The optimal transport cost deduces the Wasserstein Distance, which does not require two probability distributions to have overlapping support sets. However, when distributions are defined in different or incomparable spaces, classical OT is not applicable. For example, transporting \(s \in \mathbb{R}^2\) to \(t \in \mathbb{R}^3\) lacks a meaningful cost measurement.
Considering the well-defined distances within two distinct spaces \(\mathcal{S}\) and \(\mathcal{T}\), denoted as \(D_{\mathcal{S}}:\mathcal{S} \times \mathcal{S} \to \mathbb{R}\) and $D_{\mathcal{T}}:\mathcal{T} \times \mathcal{T} \to \mathbb{R}$, GW-OT \citep{memoli2011GromovWassersteinDistances} regards the cost of transporting \(s_i \in (\mathcal{S}, D_{\mathcal{S}})\) to \(t_j \in (\mathcal{T}, D_{\mathcal{T}})\) as the difference between the relative relationship between \(s_i\) and other elements \(s_k\) in \(\mathcal{S}\), and that between \(t_j\) and other elements \(t_l\) in \(\mathcal{T}\). Such ability to transport across incomparable spaces makes it useful in tasks such as cross-lingual alignment \citep{alvarez-melis2018GromovWassersteinAlignmentWord} and cross-domain alignment \citep{gong2022GromovWassersteinMultimodalAlignment}.

\section{Methodology}
In this section, we present the proposed graph contrastive learning framework in detail.

\subsection{THESAURUS Overview and Notations}
\begin{figure*}[t]
	\centering
	\includegraphics[width=1\linewidth]{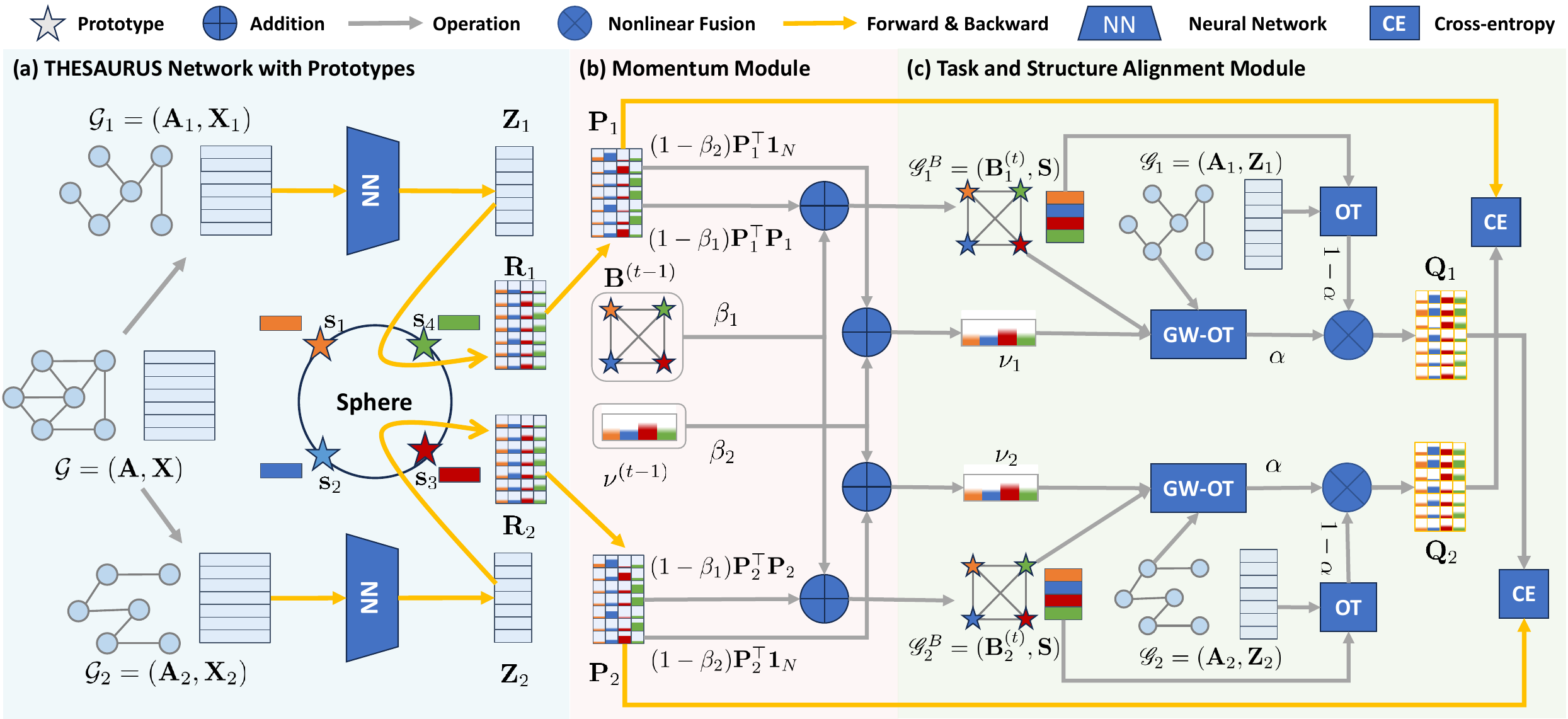}
	\caption{The illustration of our proposed THESAURUS. And the details are summarized in Algorithm \ref{alg:THESAURUS}
		in the appendix.}
	\label{fig:Arch}
\end{figure*}
The attribute graph \(\mathcal{G} = (\mathcal{V}, \mathcal{E}, \mathbf{X}) \), consisting of \(N\) nodes from \(\mathcal{V}\) and \(E\) edges from \(\mathcal{E}\), can be summarized by the tuple \(\mathcal{G} = (\mathbf{A}, \mathbf{X})\). Here, \(\mathbf{A} \in \mathbb{R}^{N \times N}\) is the (binary) adjacency matrix, and \(\mathbf{X} \in \mathbb{R}^{N \times d_0}\) is the node attribute matrix. For convenience, we will use these two graph notations interchangeably in the following sections.

Our framework is illustrated in Fig. \ref{fig:Arch}. Initially, part of edges and feature dimensions of the original graph are masked to generate two distinct (but similar) augmented views \(\mathcal{G}_1 = (\mathbf{A}_1, \mathbf{X}_1)\) and \(\mathcal{G}_2 = (\mathbf{A}_2, \mathbf{X}_2)\). Subsequently, a GCN encoder \(f_{\theta}\) \citep{kipf2017semisupervised} and an MLP projector \(f_{\omega}:\mathbb{R}^{d_1} \to \mathbb{R}^{d}\) are employed to map these views into representations \(\mathbf{Z}_1 = f_{\omega} \circ f_{\theta}(\mathbf{A}_1, \mathbf{X}_1) \in \mathbb{R}^{N\times d}\) and \(\mathbf{Z}_2 \),  where \(f_1 \circ f_2\) denotes the function composition and the whole neural network. The cosine similarities between \(\mathbf{z}_{1,i} = [\mathbf{Z}_1]_{i}\) and \(S\) semantic prototypes \(\{\mathbf{s}_i \in \mathbb{R}^{1\times d}\}_{i=1}^S\) form the context-aware representation \(\mathbf{r}_{1,i} \in \mathbb{R}^{1\times S}\) of node \(i\) in view 1, where \([\cdot]_i\) is the \(i\)-th row of matrix. Similarly, the context-aware vector \(\mathbf{r}_{2,i}\) in view 2 is obtained.

Let \(\mathbf{S}=[\mathbf{s}_1,\cdots,\mathbf{s}_S] \in\mathbb{R}^{S\times d} \) be the learnable prototype matrix, \(\mathbf{R}_1= \mathbf{Z}_1\mathbf{S}^\top \in\mathbb{R}^{N\times S}\) be the context-aware representation matrix, and \(\mathbf{B}_1\in\mathbb{R}^{S\times S}\) be the prototype graph in view 1. THESAURUS computes the (optimal) node-prototype assignment \(\mathbf{Q}_1 \in \mathbb{R}^{N \times S}\) for view 1 by solving the FGW-OT problem involving \(\mathbf{R}_1\), \(\mathbf{B}_1\), and \(\mathbf{A}_1\). Similarly, the assignment \(\mathbf{Q}_2\) is obtained. 
Once the assignments are got, we train the network \(f_{\omega} \circ f_{\theta}\) to predict \(\mathbf{Q}_2\) from \( \mathbf{Z}_1\) and vice versa. After training, the \(\mathbf{R}\) of the original graph \(\mathcal{G}\) is fed into K-means to get the final result \(\Phi \in \{0,1\}^{N\times C}\).

\subsection{Address L1: Context via Sematic Prototypes}
To capture the subtle differences between synonymous nodes, we draw inspiration from the human ability to accurately distinguish synonyms using textual context. For instance, consider the sentences \enquote{Stop being so childish!} and \enquote{She has a childlike wonder.} In the first sentence, \enquote{stop being} conveys a negative connotation, while \enquote{so} is neutral. In the second sentence, \enquote{wonder} carries a positive connotation, and \enquote{She has a} is neutral. The word co-occurrence in these sentences indicates that \enquote{childish} is associated with negative and neutral semantics (prototypes), whereas \enquote{childlike} is linked with positive and neutral semantics (prototypes). This allows us to immediately infer that \enquote{childish} has a negative meaning related to children, while \enquote{childlike} has a positive connotation, thus distinguishing these two synonyms.

Similarly, in the embedding space, we define \(S\) semantic prototypes, each representing a specific semantic category. The positional relationships between nodes and these prototypes constitute their contextual semantics. Unlike existing methods that measure the association between two nodes solely by the distance between their embedding vectors, THESAURUS uses that between \(S\)-dimensional context-aware representations --- derived from the distances between nodes and semantic prototypes --- to represent their associations. With this approach, the input for K-means is not the encoder output, but the context-aware representations \(\mathbf{R}\).

\subsection{Task and Structure Alignment Module}
To closely align with the downstream clustering task, we design the pretext task of predicting optimal clustering assignments cross views. To thoroughly exploit the cluster information in the graph structure, we align the prototype and data graph structures. These designs are integrated as the TSA module based on FGW-OT \citep{titouan2019optimal}.

\subsubsection{Address L2: Pretext Task Aligned with Clustering}
THESAURUS treats semantic prototypes as centroids providing clear semantic meanings and learns by predicting swapped clustering assignments across views.
The node-prototype assignment \(\mathbf{Q}\) is derived from context-aware representations \(\mathbf{R} = \mathbf{Z}\mathbf{S}^\top\), calculated via solving the problem
\begin{equation}
	\label{eq:OT}
	\min_{\pi \in \Pi} \Tr\left(-\mathbf{R}^\top \pi\right) - \epsilon H(\pi),
\end{equation}
where $\pi \in \mathbb{R}_{\ge0}^{N\times S} $ deduces $\mathbf{Q}$ with row-sum normalization, \(H(\pi) = -\sum_{ij} \mathbf{\pi}_{ij} \log \mathbf{\pi}_{ij}\) denotes the entropy, and \(\epsilon > 0\) controls the smoothness of the unnormalized  assignment (i.e., the coupling \(\pi\) in OT). Here \(\pi\) is constrained by the node marginal distribution \(\mu \in \mathbb{R}_{\ge0}^{N}\) and the prototype marginal distribution \(\nu \in \{\nu \in \mathbb{R}_{\ge0}^{S}\mid \sum_i\nu_i=1\}\)
\begin{equation} \label{eq:Coupling}
	\Pi = \{\pi | \pi \mathbf{1}_S = \mu, \pi^\top \mathbf{1}_N = \nu\},
\end{equation}
where \(\mathbf{1}_S\) is a \(S\)-dimensional all-one column vector. Problem \eqref{eq:OT} can be viewed as a relaxed and regularized K-means problem. K-means minimizes the sum of squared distances between data points and centroids, and we take the negative similarities $-\mathbf{R}$ as \enquote{distances} (and OT costs) here. 
Thanks to the entropy regularization, this problem can be efficiently solved by the scalable Sinkhorn \citep{cuturi2013SinkhornDistancesLightspeed}.

Feeding $\mathbf{R}_1$ of view 1 to the above procedure gives $\mathbf{Q}_1$, 
and similarly $\mathbf{Q}_2$ is got from view 2. The goal is to predict the assignment \(\mathbf{Q}_2\) of view 2 from the representation \(\mathbf{Z}_1\) of view 1, and \(\mathbf{Q}_1\) from  \(\mathbf{Z}_2\). Since \(\mathbf{Z}_1\) lacks interaction with prototypes \(\mathbf{S}\), the prediction distribution  \(\left[\mathbf{P}_1^{\tau}\right]_{n}\) for node \(n\) is built on \(\mathbf{R}_1 = \mathbf{Z}_1 \mathbf{S}^\top\) instead of \(\mathbf{Z}_1\)
\begin{equation}\label{eq:P1}
	\left[\mathbf{P}_1^{\tau}\right]_{n,s} = \frac{\exp\left(\left[\mathbf{Z}_1\right]_{n}\left[\mathbf{S}\right]_{s}^\top/\tau\right)}{\sum_{s'}\exp\left(\left[\mathbf{Z}_1\right]_{n}\left[\mathbf{S}\right]_{s'}^\top/\tau\right)},
\end{equation}
where \(\tau\) is the temperature that controls the distribution sharpness and \(\mathbf{P}_1^{1}\) is abbreviated to  \(\mathbf{P}_1\). The distributions of all nodes are stacked into \(\mathbf{P}_1^{\tau} \in \mathbb{R}_{\ge0}^{N \times S}\). Similarly, $\mathbf{P}_2^{\tau}$ is obtained. The overall training loss is then computed as the averaged cross-entropy 
\begin{equation}
	\label{eq:Loss}
	\fontsize{7.2pt}{10pt}\selectfont
	\mathcal{L} = -\frac{1}{2N} \sum_{n=1}^{N} \sum_{s=1}^{S} \left( \left[\mathbf{Q}_1\right]_{n,s} \log \left[\mathbf{P}_2^{\tau}\right]_{n,s} + \left[\mathbf{Q}_2\right]_{n,s} \log \left[\mathbf{P}_1^{\tau}\right]_{n,s} \right)
\end{equation}


\subsubsection{Address L3: Structure Alignment via GW-OT}
The above node-prototype clustering assignment is derived from the relationships between node embeddings \(\mathbf{Z}\) and prototypes \(\mathbf{S}\). Like prior methods, this approach does not explicitly extract structure cluster information. To fill this gap, we propose deriving the optimal assignment from the structure and using this assignment as a prediction target for \(\mathbf{Z}\). One effective method to assign nodes to different semantic prototypes based on the structures is to perform GW-OT between \(\mathbf{A}\) and an isolated graph \(\mathbf{I}_{S}\) \citep{xu2019GromovWassersteinLearningGraph}. In this isolated graph, each node represents a cluster with no inter-cluster edges. Such approach adheres to cluster definition but ignores inter-cluster relationships, which is unrealistic. Therefore, we replace the isolated graph with a complete prototype graph \(\mathbf{B} \in \mathbb{R}_{\ge0}^{S \times S} \), which is constructed as
\begin{equation}
	\mathbf{B} = \mathbf{P}^{\top}\mathbf{P}.
\end{equation}

GW-OT is formally defined in Def. \ref{def:GW-OT-Graph}. For an attribute graph \(\mathcal{G}=(\mathcal{V}, \mathcal{E}, \mathbf{X})\), each node \(v_i\) contains observable attribute \(x_i = \left[\mathbf{X}\right]_{i,:} \in \Omega_x \subset \mathbb{R}^d\) and implicit structure embedding \(s_i \in \Omega_s\). Although \(s_i\) is not known, the pairwise relationship \(\mathbf{C} \in \mathbb{R}_{\ge 0}^{N \times N}\) determined by the metric \(D_{\Omega_s}: \Omega_s \times \Omega_s \to \mathbb{R}_{\ge 0}\) on the space \(\Omega_s\) is given by the adjacency matrix \(\mathbf{A}\), the Laplacian \(\mathbf{L}\), or the pairwise shortest path matrix \citep{chowdhury2019GromovWassersteinDistance}. To use OT, the probability measure on the space \(\left(\mathcal{V}, D_{\Omega_s}\right)\) or equivalently \(\left(\mathcal{V}, \mathbf{C}\right)\) must be defined. Denote the importance of $N$ nodes by the histogram 
\begin{equation}
	h \in \mathcal{H}_N=\left\{ h \mid h \in \mathbb{R}_{>0}^{N}, \sum_{i=1}^{N} h_i = 1 \right \}.
\end{equation}
Then this space has a measure \(\mu = \sum_{i} h_i \delta_{(s_i)}\), where \(\delta_{(s_i)}\) denotes the Dirac delta function at \(s_i\).
\begin{defn}{}
	\label{def:GW-OT-Graph}
	Let \(\left(\mathcal{V}_{1}, \mathbf{C}_{1}, \mu\right)\) and \(\left(\mathcal{V}_{2}, \mathbf{C}_{2}, \nu\right)\) be Metric-Measure (MM) spaces defined on \(\mathcal{G}_{1} = (\mathbf{C}_{1}, \emptyset)\) and \(\mathcal{G}_{2} = (\mathbf{C}_{2}, \emptyset)\), respectively.
	$\mu = \sum_{i} h_{i}^{(1)} \delta_{(s_{i})},h^{(1)} \in \mathcal{H}_{N_1}
	$ and 
	$\nu = \sum_{i} h_{i}^{(2)} \delta_{(s_{i})},h^{(2)} \in \mathcal{H}_{N_2} $ are the probability measures on $\mathcal{V}_1$ and $\mathcal{V}_2$, separately. The Gromov-Wasserstein distance \(GW_{p}(\mathcal{G}_{1}, \mathcal{G}_{2}) \) between these two measures is given by
	\begin{equation}
		\inf_{\pi \in \Pi} \sum_{i,k=1}^{N_{1}} \sum_{j,l=1}^{N_{2}} \left( \left[\mathbf{C}_{1}\right]_{i,k} - \left[\mathbf{C}_{2}\right]_{j,l} \right)^{p} \pi_{i,j} \pi_{k,l},
	\end{equation}
	where \(\Pi= \left\{ \pi \in \mathbb{R}_{\ge 0}^{N_{1} \times N_{2}} \mid  \pi \mathbf{1}_{N_2} = h^{(1)}, \pi^{\top}  \mathbf{1}_{N_1}= h^{(2)} \right\}\).
\end{defn}
We can utilize the \(GW_{1}\) OT between the non-attribute data graph \((\mathbf{A},\emptyset)\) and prototype graph \((\mathbf{B},\emptyset)\) to get the optimal node-prototype assignment \(\mathbf{Q}\). It is row-normalized from the solution \(\pi\) of following OT problem
\begin{equation} \label{eq:GW-OT-AB}
	\min_{\pi \in \Pi} \sum_{i,k=1}^{S} \sum_{j,l=1}^{N} \left| \left[\mathbf{A}\right]_{i,k} - \left[\mathbf{B}\right]_{j,l} \right| \pi_{i,j} \pi_{k,l},
\end{equation}
where \(\mu\) is the uniform node marginal distribution and \(\nu\) is the current prototype marginal. After the structure-induced assignments \(\mathbf{Q}_1, \mathbf{Q}_2\) of two views \(\mathcal{G}_1,\mathcal{G}_2\) are separately got via Eq. \eqref{eq:GW-OT-AB}, they can be used with the loss Eq. \eqref{eq:Loss}.

\subsubsection{Fused Clustering Assignment via FGW-OT}
The above introduce two kinds of \enquote{clustering} assignments respectively acquired from the context-aware node representation $\mathbf{R}$ and the graph structure $\mathbf{A}$. The assignment from $\mathbf{R}$ focuses on attribute information, while that from $\mathbf{A}$ emphasizes structural information. We fuse them with FGW-OT for more comprehensive graph mining. We build \(\mathscr{G} = (\mathbf{A}, \mathbf{Z})\) with the embeddings \(\mathbf{Z}\) as node attributes. And we add prototypes \(\mathbf{S} \in \mathbb{R}^{S \times d}\) as attributes to the prototype graph \(\mathbf{B}\), resulting in \(\mathscr{G}^B = (\mathbf{B}, \mathbf{S})\). The optimal transport between \(\mathscr{G}^B\) and \(\mathscr{G}\) encapsulate both attribute and structure cluster information, and can be achieved via FGW-OT defined below.
\begin{defn}{}
	\label{def:FGW-OT}
	Let \(\left(\mathcal{V}_{1}, \mathbf{C}_{1}, \mu\right)\) and \(\left(\mathcal{V}_{2}, \mathbf{C}_{2}, \nu\right)\) be MM-spaces on \(\mathcal{G}_{1} = (\mathbf{C}_{1}, \mathbf{X}_{1})\) with measure 
	\(\mu = \sum_{i} h_{i}^{(1)} \delta_{(s_{i}, x_{i})},h^{(1)} \in \mathcal{H}_{N_1} \)  
	and 
	on \(\mathcal{G}_{2} = (\mathbf{C}_{2}, \mathbf{X}_{2})\) with measure 
	\(\nu = \sum_{i} h_{i}^{(2)} \delta_{(s_{i}, x_{i})}, h^{(2)} \in \mathcal{H}_{N_2} \), respectively. The Fused Gromov-Wasserstein distance \(FGW_{p,\alpha}(\mathcal{G}_{1}, \mathcal{G}_{2})\) is
	\begin{align} \label{eq:FGW}
		\inf_{\pi \in \Pi} \Bigg\{ \sum_{i,k=1}^{N_{1}} \sum_{j,l=1}^{N_{2}} \bigg[ (1 - \alpha) D_{\Omega_{x}}(x_{1,i}, x_{2,j}) & \notag \\
		+ \alpha |\mathbf{C}_{1}(i,k) - \mathbf{C}_{2}(j,l)|^{p} \bigg] \pi_{i,j} \pi_{k,l} \Bigg\}^{\frac{1}{p}}  - \epsilon H(\pi) &
	\end{align}
	where \(\Pi= \left\{ \pi \in \mathbb{R}_{\ge 0}^{N_{1} \times N_{2}} \mid  \pi \mathbf{1}_{N_2} = h^{(1)}, \pi^{\top}  \mathbf{1}_{N_1}= h^{(2)} \right\}\);
	\(H(\pi) = -\sum_{ij} \mathbf{\pi}_{ij} \log \mathbf{\pi}_{ij}\) is the coupling entropy and $\epsilon$ weights this regularization;
	\(x_{1,i}, x_{2,j} \in \Omega_{x} \subset \mathbb{R}^{d}\) are the attributes of nodes \(v_{i}^{(1)} \in \mathcal{V}_{1}\) and \(v_{j}^{(2)} \in \mathcal{V}_{2}\), respectively.
\end{defn}
The optimal coupling of \(FGW_{1,\alpha}(\mathscr{G}, \mathscr{G}^{B})\) encapsulates the information in $\mathbf{R}$ and $\mathbf{A}$,where $\alpha$ balances the contribution of these two parts. In view 1, we can construct \( \mathscr{G}_1=(\mathbf{A}_1,\mathbf{Z}_1)\) and \(  \mathscr{G}_1^B=(\mathbf{B}_1, \mathbf{S}) \), and then the assignment $\mathbf{Q}_1$ of 
\(FGW_{1,\alpha}(\mathscr{G}_1, \mathscr{G}_1^B)\) is obtained via normalizing the optimal coupling
$\pi_1$ at each row $n\in \{1,2,\dots,N\}$
\begin{equation} \label{eq:pi2Q}
	[\mathbf{Q}_1]_{n,s} =\frac{[\pi_1]_{n,s}}{ \sum_{s'=1}^S [\pi_1]_{n,s'}}.
\end{equation}
Similarly, \(\mathbf{Q}_2\) is got from view 2. Finally, these fused assignments are used to guide the training with Eq. \eqref{eq:Loss}.

\subsection{Prototype Momentum Module}
Prototype graph and marginal should adapt to the data. So momentum update is adopted here. Let $\mathbf{B}^{(t-1)}$ be the prototype graph of last forward step \(t-1\). The step $t$ has
\begin{equation}
	\mathbf{B}^{(t)} = \beta_1 \mathbf{B}^{(t-1)} + (1-\beta_1) {\mathbf{P} }^{\top}\mathbf{P}.
\end{equation}
To reflect the common graph homophily, $\mathbf{B}$ is initialized as an identity matrix and \(\beta_1 \in [0, 1]\) is set to a high value. 

Meanwhile, the logits $\mathbf{P}$, e.g, Eq. \eqref{eq:P1}, are used to update prototype marginal, which is initialized as an uniform distribution \(\nu^{(0)} =\mathbf{1}_S / S\) to reflect the presumed even cluster size before data exposure. 
\begin{equation}
	\nu^{(t)} = \beta_2 \nu^{(t-1)} + (1-\beta_2) (\mathbf{P}^{\top} \mathbf{1}_N)
\end{equation}

\section{Experiments}

\begin{table*}[htb]
	\centering 	
	\fontsize{9pt}{9pt}\selectfont 	
	\setlength{\tabcolsep}{1mm}
	\begin{tabular}{ccccccccccccccc}
		\toprule 
		Datasets & Metrics & K-means & DEC & SDCN & GRACE & DAEGC & DFCN & DCRN & HSAN & S$^3$GC & SCGC & Dink-Net{*} & Dink-Net & Ours\tabularnewline
		\midrule 
		\multirow{4}{*}{Cora} & ACC & 33.80 & 46.50 & 35.60 & 73.90 & 70.43 & 36.33 & 61.93 & 77.21 & 74.21 & 73.88 & 75.55 & 78.21 & \textbf{80.72}\tabularnewline
		& NMI & 14.98 & 23.54 & 14.28 & 57.10 & 52.89 & 19.36 & 45.13 & 59.56 & 58.80 & 56.10 & 60.03 & 62.48 & \textbf{62.92}\tabularnewline
		& ARI & 8.60 & 15.13 & 7.78 & 52.70 & 49.63 & 4.67 & 33.15 & 57.93 & 54.43 & 51.79 & 54.56 & 61.48 & \textbf{63.61}\tabularnewline
		& F1 & 30.26 & 39.23 & 24.37 & 72.50 & 68.27 & 26.16 & 49.50 & 75.13 & 72.10 & 70.81 & 71.31 & 72.85 & \textbf{78.44}\tabularnewline
		\midrule 
		\multirow{4}{*}{Citeseer} & ACC & 39.32 & 46.51 & 65.96 & 63.10 & 64.54 & 69.50 & 69.86 & 71.05 & 68.81 & 71.02 & 69.34 & 69.91 & \textbf{71.99}\tabularnewline
		& NMI & 16.94 & 23.54 & 38.71 & 39.91 & 36.41 & 43.90 & 44.86 & 45.62 & 44.11 & 45.25 & 44.36 & 45.29 & \textbf{47.37}\tabularnewline
		& ARI & 13.43 & 15.13 & 40.17 & 37.70 & 37.78 & 45.50 & 45.64 & 48.22 & 44.80 & 46.29 & 45.65 & 46.29 & \textbf{48.99}\tabularnewline
		& F1 & 36.08 & 39.23 & 63.62 & 60.30 & 62.24 & 64.30 & 64.83 & 64.52 & 64.30 & 64.80 & 65.54 & 65.79 & \textbf{66.42}\tabularnewline
		\midrule 
		\multirow{4}{*}{Pubmed} & ACC & 59.83 & 60.14 & 64.20 & 63.72 & 68.73 & 68.89 & 69.87 & \multirow{4}{*}{OOM} & 71.31 & 45.12 & 67.32 & 67.51 & \textbf{79.64}\tabularnewline
		& NMI & 31.05 & 22.14 & 22.87 & 30.86 & 28.26 & 31.43 & 32.20 &  & 33.35 & 7.04 & 32.49 & 33.01 & \textbf{41.43}\tabularnewline
		& ARI & 28.10 & 19.55 & 22.30 & 27.61 & 29.84 & 30.64 & 31.41 &  & 34.52 & 7.04 & 29.95 & 30.44 & \textbf{48.25}\tabularnewline
		& F1 & 58.88 & 61.45 & 65.01 & 62.85 & 68.23 & 68.10 & 68.94 &  & 70.33 & 44.54 & 67.12 & 67.35 & \textbf{79.00}\tabularnewline
		\midrule 
		\multirow{4}{*}{A-Photo} & ACC & 27.22 & 47.22 & 53.44 & 67.66 & 75.96 & 76.82 & 79.94 & 77.02 & 75.15 & 77.48 & 77.19 & 80.71 & \textbf{84.42}\tabularnewline
		& NMI & 13.23 & 37.35 & 44.85 & 53.46 & 65.25 & 66.23 & 73.70 & 67.54 & 59.78 & 67.67 & 68.94 & 70.50 & \textbf{74.99}\tabularnewline
		& ARI & 5.50 & 18.59 & 31.21 & 42.74 & 58.12 & 58.28 & 63.69 & 58.05 & 56.13 & 58.48 & 60.20 & 66.54 & \textbf{72.01}\tabularnewline
		& F1 & 23.96 & 46.71 & 50.66 & 60.32 & 69.87 & 71.25 & 73.82 & 72.60 & 72.85 & 72.22 & 71.23 & 73.09 & \textbf{76.49}\tabularnewline
		\midrule
		\multirow{4}{*}{CoraFull} & ACC & 26.27  & 31.92 & 26.67 & 32.38 & 34.35 & 37.51 & 38.80 & \multirow{4}{*}{OOM} & 36.46 & 41.89 & 38.51 & 39.45 & \textbf{42.94}\tabularnewline
		& NMI & 34.68  & 41.31 & 37.83 & 50.42 & 49.16 & 51.30 & 51.91 &  & 52.82 & 53.21 & 53.39 & 54.41 & \textbf{55.83}\tabularnewline
		& ARI & 9.35  & 16.89 & 22.60 & 20.64 & 22.60 & 24.46 & 25.25 &  & 24.78 & 24.23 & 26.53 & 27.45 & \textbf{30.07}\tabularnewline
		& F1 & 22.57 & 27.77 & 22.14 & 27.82 & 26.96 & 31.22 & 31.68 &  & 29.78 & 32.98 & 30.90 & 31.95 & \textbf{37.01}\tabularnewline
		\bottomrule
	\end{tabular}
	\caption{Clustering performance (\%). The best result is in bold. Dink-Net{*} denotes Dink-Net-NoFT.}
	\label{tab:MainResults}
\end{table*}

\subsection{Experimental Setup}
To evaluate the performance of THESAURUS, we run the proposed  method on nine attribute graph datasets, including Cora, Citeseer, Pubmed, Amazon-Photo (A-Photo), CoraFull, ACM, DBLP, UAT, and Wiki. The baselines are K-means, 
DEC,
GRACE \citep{zhu2020DeepGraphContrastive},
SDCN,
DFCN,
DCRN,
S$^3$GC,
SCGC,
HSAN, and
Dink-Net.

Our evaluation protocol follows that of the previous SOTA Dink-Net \citep{liu2023DinkNetNeuralClustering}. Besides Normalized Mutual Information (NMI) and Adjusted Rand Index (ARI), the metrics include Accuracy (ACC) and the Macro-F1 score (F1), computed after mapping the clusters to the ground-truth classes with linear assignment \citep{lovasz1986MatchingTheoryNorthHolland,crouse2016Implementing2DRectangular}. The F1 score, defined as the harmonic mean of precision and recall, balances the effects of false positives and false negatives. Meanwhile, ARI quantifies the number of true positive and true negative pairs and normalizes these values to ensure that the assessment is not influenced by variations in cluster sizes. Therefore, F1 and ARI are more effective than ACC and NMI on imbalanced data.

\subsection{Overall Performance}
Part of the results are summarized in Table \ref{tab:MainResults}, with OOM indicating out-of-memory failures on one RTX 4090 GPU. And the rest results are presented in Tables \ref{tab:ExtraResutls1} and \ref{tab:ExtraResutls2} 
in the appendix. These results demonstrate that the proposed THESAURUS significantly outperforms existing methods across all datasets. And several key observations can be made.

\textbf{The contextual information from semantic prototypes is important.} Existing methods do not achieve sufficient cluster separability and are affected by the Uniform Effect and Cluster Assimilation. This results in suboptimal performance on clusters of varying sizes, particularly minority clusters, adversely impacting F1 and ARI. In contrast, THESAURUS, utilizing semantic prototype contexts, achieves better distinction between synonymous nodes, leading to higher cluster separability. This effectively mitigates Uniform Effect and Cluster Assimilation, evidenced by THESAURUS's F1 and ARI significantly surpassing existing methods.
\textbf{Pretext task aligned with clustering is vital.} Contrastive clustering methods like S$^3$GC and HSAN outperform DEC-style methods such as SDCN and DFCN, likely due to the implicit but insufficient alignment between InfoNCE and (spectral) clustering \citep{haochen2021ProvableGuaranteesSelfsupervised, tan2023ContrastiveLearningSpectral}. DinkNet, which explicitly optimize towards clustering during finetune, is better aligned with clustering tasks, thus surpassing other baselines. Furthermore, THESAURUS representations are learned towards high cluster separability from start to finish, and thus far outperform all other methods.
\textbf{The cluster information in structure matters.} Methods utilizing no graph structures, such as K-means and DEC, are unsuitable for graph clustering. Methods like HSAN and SCGC, which inject structure information with supervision signals, generally surpass those that only leverage structure through graph filters, such as SDCN and DAEGC. Furthermore, THESAURUS, which exhaustively exploits structural cluster information, surpasses all other methods.
\begin{figure}[h]
	\centering
	\begin{subfigure}{0.495\columnwidth}
		\includegraphics[width=1\linewidth]{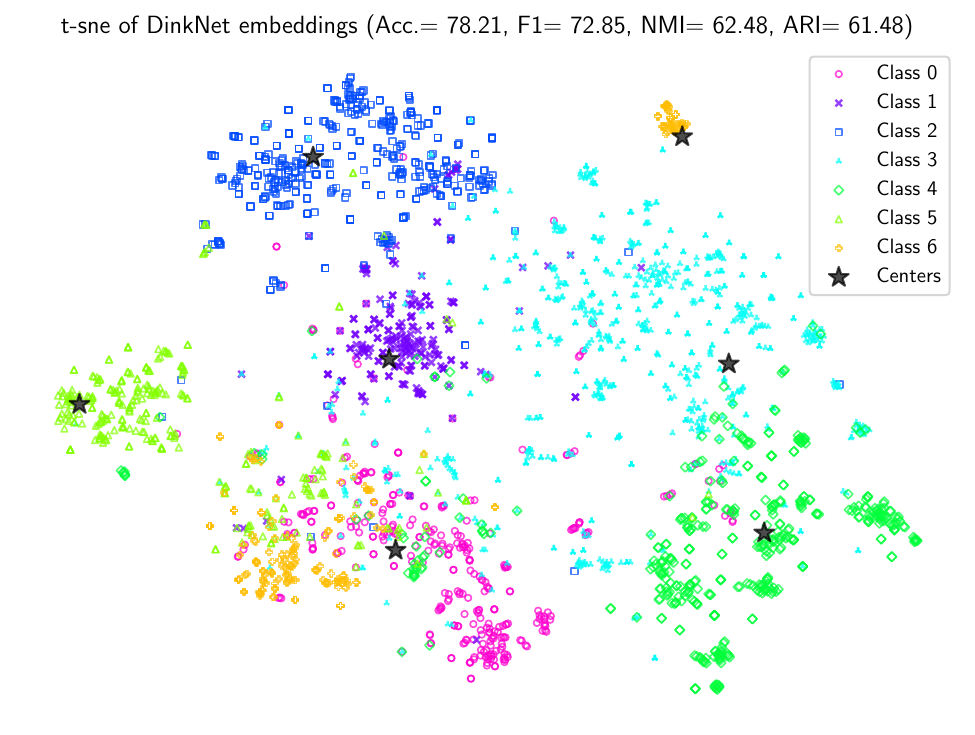}
		\caption{Dink-Net on Cora}
		\label{fig:TSNE_DinkNet_Cora}
	\end{subfigure}
	\begin{subfigure}{0.495\columnwidth}
		\includegraphics[width=1\linewidth]{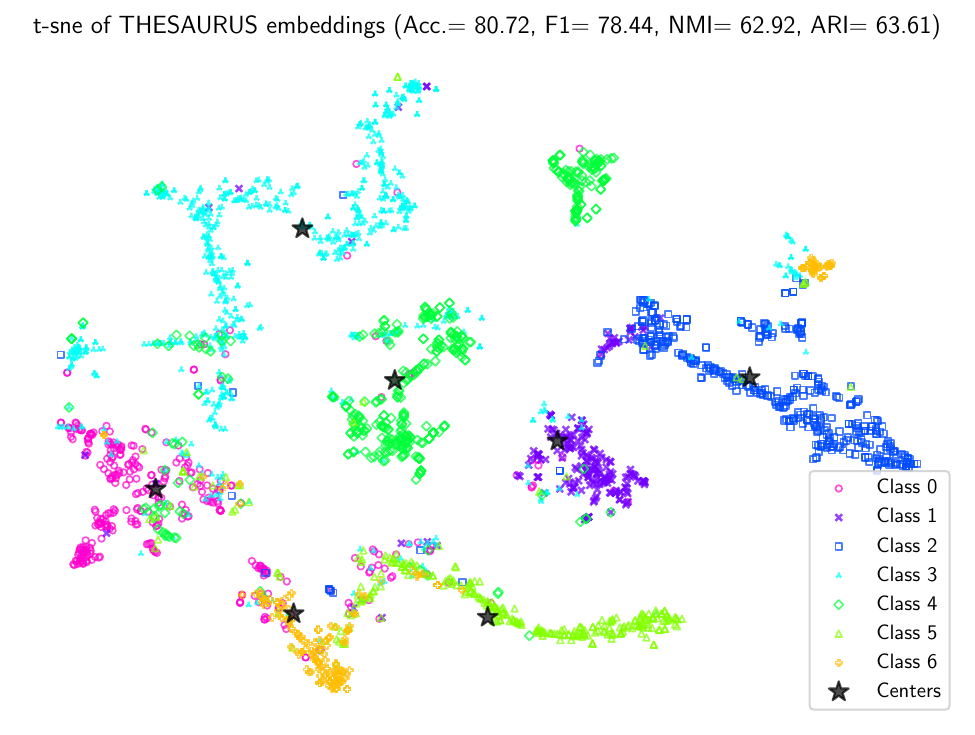}
		\caption{THESAURUS on Cora}
		\label{fig:TSNE_THESAURUS_Cora}
	\end{subfigure}
	\caption{The visualization of Dink-Net and THESAURUS}
	\label{fig:Visual2D}
\end{figure}

\subsection{Class-wise Performance \& Visualization}
\begin{figure}[t]
	\includegraphics[width=1\linewidth]{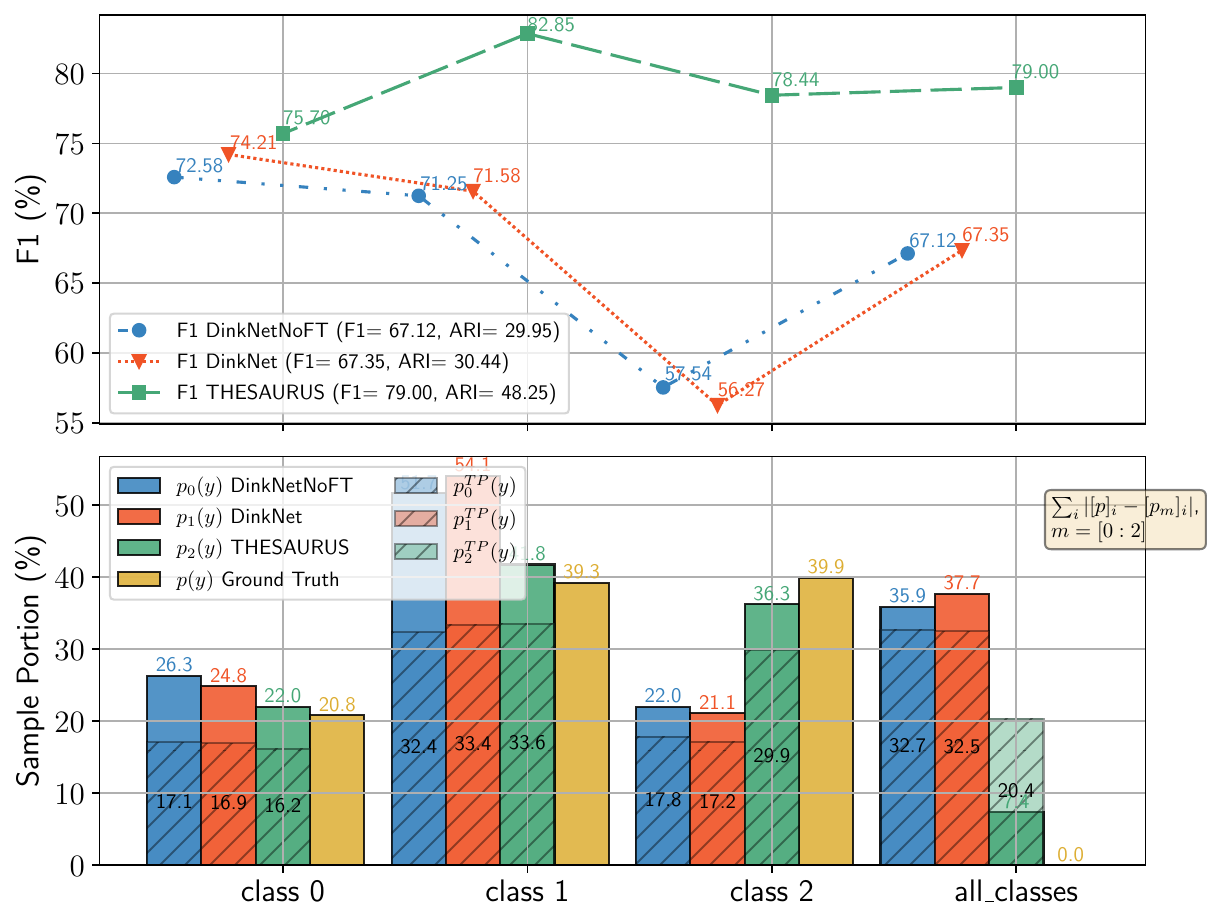}
	\caption{Dink-Net and THESAURUS on Pubmed. The \textbf{top} figure illustrates the F1 scores for each category, as well as the Macro-F1. The \textbf{bottom} shows the distribution of labels predicted by Dink-Net and THESAURUS, along with the ground-truth labels. It also presents the distribution of predicted labels for true-positive (TP) samples, denoted as \(p_{i}^{TP}(y), i\in  \{0,1,2\}\). The final set of bars shows the differences between the predicted and ground-truth distributions.}
	\label{fig:DinkNetvsTHESAURUS_pubmed}
\end{figure}

While Dink-Net's finetune step mitigates the Uniform Effect in the majority cluster (cluster 3), it fails to address Cluster Assimilation in the minority cluster (cluster 6), as indicated by the experiments presented in Section Introduction. This section evaluates the class-wise performance of Dink-Net and THESAURUS, demonstrating that THESAURUS achieves high cluster separability and addresses the failure cases of  Dink-Net. Fig. \ref{fig:Visual2D} visualizes the final representations to cluster and the centroids on Cora using t-SNE \citep{maaten2008VisualizingDataUsing}. Fig. \ref{fig:DinkNetvsTHESAURUS_pubmed} displays the class-wise performance on Pubmed, in a style like Fig. \ref{fig:DinkNetFintuneEffect-F1}.

Overall, the predicted label distribution of THESAURUS exhibits a lower deviation from the ground-truth distribution than that of DinkNet, as shown in Fig. \ref{fig:DinkNetFintuneEffect-F1} and Fig. \ref{fig:DinkNetvsTHESAURUS_pubmed}. This suggests fewer mis-clustered nodes and corresponds with the higher cluster separability observed in THESAURUS's representation space in Fig. \ref{fig:Visual2D}, resulting in a Macro-F1 score up to 5.5 percentage points higher than Dink-Net on Cora and 11.65 points higher on Pubmed. 

\paragraph{Uniform Effect} Fig. \ref{fig:TSNE_DinkNet_Cora} (right part) shows that Dink-Net does not separate clusters 3 and 4 as well as THESAURUS in Fig. \ref{fig:TSNE_THESAURUS_Cora} (top-left). So although Dink-Net reduces the Uniform Effect in cluster 3, the F1 score for cluster 4 is significantly lower than THESAURUS, as shown by Fig. \ref{fig:DinkNetFintuneEffect-F1}. Besides, Dink-Net's slightly higher F1 score for cluster 3 comes at the cost of many false positives, which severely compromises the performance of other clusters (e.g., clusters 4 and 0). In contrast, THESAURUS doesn't favor the majority cluster and performs well overall. 

\paragraph{Cluster Assimilation} Fig. \ref{fig:TSNE_DinkNet_Cora} (bottom-left) shows Dink-Net mixing clusters 0, 5, and 6, leading to a low F1 score of about 34\% for cluster 6 due to samples being merged into other clusters. Fig. \ref{fig:TSNE_THESAURUS_Cora} shows that THESAURUS effectively separates these clusters, significantly mitigating Cluster Assimilation, as evidenced by a 27.29 percentage point higher F1 score for cluster 6 compared to Dink-Net (see Fig. \ref{fig:DinkNetFintuneEffect-F1}).

\subsection{Ablation Study}
To validate the effectiveness of the complete prototype graph, we set the prototype graph \(\mathbf{B}\) as isolated graph \(\mathbf{I}_S\) (w/o \(\mathbf{B}\)). To test the impact of structural information extraction, we replace \(\mathbf{A}_1\) and \(\mathbf{A}_2\) in FGW-OT with \(\mathbf{I}_N\) (w/o \(\mathbf{A}\)). Additionally, to assess the effectiveness of the proposed momentum module, we set the momentum to 1 (fixed \(\nu\) \& \(\mathbf{B}\)). The results in Table \ref{tab:Ablation} indicate that the complete prototype graph outperforms the isolated one, structural information extraction is effective, and momentum updates for the prototype graph and marginal is useful.

\begin{table}
	\centering
	\fontsize{9pt}{9pt}\selectfont 	
	\begin{tabular}{cccccc}
		\toprule 
		Datasets & Metrics & w/o $\mathbf{B}$ & w/o $\mathbf{A}$ & fixed \(\nu\) \& \(\mathbf{B}\)   & Ours\tabularnewline
		\midrule 
		\multirow{4}{*}{Citeseer} & ACC & 70.27 & 71.54 & 71.24 & \textbf{71.99}\tabularnewline
		& NMI & 46.41 & 46.74 & 46.37 & \textbf{47.37}\tabularnewline
		& ARI & 47.82 & 47.92 & 47.57 & \textbf{48.99}\tabularnewline
		& F1 & 65.02 & 65.75 & 65.68 & \textbf{66.42}\tabularnewline
		\midrule 
		\multirow{4}{*}{Pubmed} & ACC & 75.73 & 78.35 & 75.84 & \textbf{79.64}\tabularnewline
		& NMI & 35.44 & 39.19 & 35.50 & \textbf{41.43}\tabularnewline
		& ARI & 39.25 & 45.47 & 40.62 & \textbf{48.25}\tabularnewline
		& F1 & 75.27 & 77.81 & 75.39 & \textbf{79.00 }\tabularnewline
		\bottomrule
	\end{tabular}
	\caption{THESAURUS ablation. The best is in bold.}
	\label{tab:Ablation}
\end{table}

\section{Conclusion}
This work identifies challenges in prior deep graph clustering methods, particularly the Uniform Effect and Cluster Assimilation issues, which arise due to low cluster separability in the learned embedding space. To address these challenges, we propose a novel contrastive graph learning framework, THESAURUS. Our method 1) utilizes semantic prototypes to provide contextual information crucial for distinguishing similar nodes from different classes, 2) leverages a pretext task well-aligned with the downstream clustering task for better feature transferability, 3) takes GW-OT to thoroughly exploit the cluster information in graph structure, and 4) employs a momentum module for data adaptability. To achieve comprehensive information mining, cross-view alignments used in the pretext task are acquired via  FGW-OT. This approach integrates designs 2 and 3 organically, providing an unified TSA module for better alignment. Experimental results strongly validate the effectiveness and superiority of THESAURUS compared to existing methods, showcasing its enhanced capability in achieving more accurate and efficient alignments.

\section*{Acknowledgements}
The research is supported by the National Key Research and Development Program of China (2023YFB2703700), the National Natural Science Foundation of China (62176269), and the Guangzhou Science and Technology Program (2023A04J0314).

\section*{Ethics Statement}
This research utilizes publicly available datasets and comparison methods, all of which are based on open-source code. No human participants or private data are involved in this study. All datasets used have been anonymized, and ethical guidelines regarding data usage have been strictly followed. We ensure that the methods used are transparent.

\bibliography{MyLibrary}

\clearpage
\appendix

\section{Algorithm of THESAURUS}

The pseudo-code for THESAURUS is given in Algorithm \ref{alg:THESAURUS}. Please also refer to the schematic Fig. \ref{fig:Arch} in the main content for a quick and comprehensive understanding of the details of THESAURUS.
\begin{algorithm}[H]
	\caption{Pseudocode of THESAURUS}
	\label{alg:THESAURUS}
	\textbf{Input}: The graph \(\mathcal{G}=(\mathbf{A},\mathbf{X})  \) with \(E\) edges and \(d_0\) attribute dimensions; the number of clusters \(C\); training epochs $T$; momentum weights \(\beta_1\) and \(\beta_2\);  the edge drop rate \(pe\) and feature drop rate \(px\) of data augmentation module\\
	\textbf{Parameter}: the trade-off weight \(\alpha\) of FGW-OT; Softmax temperature \(\tau\); the number of prototypes \(S\) \\
	\textbf{Output}: Clustering assignment \(\Phi\)
	\begin{algorithmic}[1]
		\STATE Initialize current epoch $t=1$, neural network \(f_{\omega} \circ f_{\theta}\), prototypes \(\mathbf{S}\), prototype graph $\mathbf{B}^{(0)}=\mathbf{I}_S$, and uniform prototype marginal \(\nu^{(0)} = \mathbf{1}_{S}/S\)\\
		\WHILE {\(t<T\)}
		\STATE \textit{\textcolor{blue}{//* Generate two augmented views} } \\
		\STATE Randomly drop \(pe\cdot E\) edges and \(px \cdot d_0\) feature dimensions to get
		\(\mathcal{G}_1=(\mathbf{A}_1,\mathbf{X}_1) \) and \(\mathcal{G}_2=(\mathbf{A}_2,\mathbf{X}_2) \)
		\STATE \textit{\textcolor{blue}{//* Encode two views with  \(f_{\omega} \circ f_{\theta}\)} } \\
		\STATE Encode \(\mathcal{G}_1\) and \(\mathcal{G}_2\) with \(f_{\omega} \circ f_{\theta}\) into \(\mathbf{Z}_1\) and  \(\mathbf{Z}_2\) \\
		
		\STATE \textit{\textcolor{blue}{//* Process view 1 }} \\
		\STATE \(\mathbf{R}_1=\mathbf{Z}_1 \mathbf{S}^{\top}\),  compute \(\mathbf{P}_1\) and \(\mathbf{P}_1^{\tau}\) with Eq. \eqref{eq:P1}
		\STATE \(\mathbf{B}_1^{(t)}=\beta_1 \mathbf{B}^{(t-1)}+(1-\beta_1) { \mathbf{P}_1}^{\top} \mathbf{P}_1\)
		\STATE \(  \nu_1^{(t)}=\beta_2\nu^{(t-1)}+(1-\beta_2)(\mathbf{P}_1^\top\mathbf{1}_N)\)
		
		\STATE \(\mathbf{B}^{(t-1)} \gets \mathbf{B}_1^{(t)}\); \(\nu^{(t-1)}\gets  \nu_1^{(t)} \)
		
		\STATE \textit{\textcolor{blue}{//* Process view 2 }} \\
		\STATE \(\mathbf{R}_2=\mathbf{Z}_2 \mathbf{S}^{\top}\),  compute \(\mathbf{P}_2\) and 
		\(\mathbf{P}_2^{\tau}\) like Eq. \eqref{eq:P1}
		\STATE \(\mathbf{B}_2^{(t)}=\beta_1 \mathbf{B}^{(t-1)}+(1-\beta_1) { \mathbf{P}_2}^{\top} \mathbf{P}_2\)
		\STATE \( \nu_2^{(t)}=\beta_2\nu^{(t-1)}+(1-\beta_2)(\mathbf{P}_2^\top\mathbf{1}_N)\)
		
		\STATE \(\mathbf{B}^{(t)} \gets \mathbf{B}_2^{(t)}\); \(\nu^{(t)}\gets  \nu_2^{(t)} \)
		
		\STATE \textit{\textcolor{blue}{//* Prepare attribute graphs for FGW-OT}} \\
		
		\STATE \(\mathbf{B}_1 \gets \mathbf{B}_1^{(t)} \); \(\mathbf{B}_2 \gets \mathbf{B}_2^{(t)} \); \( \nu_1 \gets \nu_1^{(t)} \); \( \nu_2 \gets \nu_2^{(t)} \) 
		\STATE Construct graphs \( \mathscr{G}_1=(\mathbf{A}_1,\mathbf{Z}_1)\) and \(  \mathscr{G}_1^B=(\mathbf{B}_1, \mathbf{S}) \)
		\STATE Construct graphs \( \mathscr{G}_2=(\mathbf{A}_2,\mathbf{Z}_2)\) and 
		\(  \mathscr{G}_2^B=(\mathbf{B}_2, \mathbf{S}) \)
		
		\STATE Set \( \nu_1\) as the measure on \(\mathscr{G}_1^B\) and   \( \nu_2\) on \(\mathscr{G}_2^B\) \\
		\STATE Set uniform measures \(\mu_1\) for \(\mathscr{G}_1\) and  \(\mu_2\) for \(\mathscr{G}_2\)
		
		\STATE \textit{\textcolor{blue}{//* Compute the optimal couplings of FGW-OT}} \\
		
		\STATE Get the optimal coupling \(\pi_1\) of \(FGW_{1,\alpha}(\mathscr{G}_1, \mathscr{G}^{B}_1)\)
		\STATE Get the optimal coupling \(\pi_2\) of \(FGW_{1,\alpha}(\mathscr{G}_2, \mathscr{G}^{B}_2)\)
		
		\STATE \textit{\textcolor{blue}{//* Compute the cross-view loss and backward}} \\
		
		\STATE Get \(\mathbf{Q}_1, \mathbf{Q}_2\) with Eq. \eqref{eq:pi2Q}
		\STATE Compute the loss Eq. \eqref{eq:Loss} with \(\mathbf{Q}_1, \mathbf{Q}_2,\mathbf{P}_1^{\tau}, \mathbf{Q}_2^{\tau}\)
		\STATE Backward and update \(f_{\omega} \circ f_{\theta}\) and \(\mathbf{S}\)
		\STATE $t\gets t+1$
		\ENDWHILE 
		
		\STATE \textit{\textcolor{blue}{//* Inference with original data, and final clustering}} \\ 
		\STATE Encode \(\mathcal{G}=(\mathbf{A},\mathbf{X})\) into \(\mathbf{Z}\) with the trained  \(f_{\omega} \circ f_{\theta}\) 
		\STATE Feed \(\mathbf{R}=\mathbf{Z} \mathbf{S}^{\top}\) into K-means to get \(\Phi\) over \(C\) classes
		
	\end{algorithmic}
\end{algorithm}

\section{Implementation Details}

\subsection{Datasets}
We use nine public datasets
\begin{itemize}
	
	\item Cora, Citeseer, and Pubmed from \cite{sen2008CollectiveClassificationNetwork}
	
	\item Amazon-photo (A-Photo) from \cite{shchur2019PitfallsGraphNeural}
	
	\item CoraFull from \cite{bojchevski2018DeepGaussianEmbedding}
	
	\item  ACM and DBLP from \cite{bo2020StructuralDeepClustering}
	
	\item UAT from \cite{mrabah2023RethinkingGraphAutoEncoder}
	
	\item Wiki from \cite{cao2016DeepNeuralNetworks}.
\end{itemize}

Cora, Citeseer, Pubmed, CoraFull, and Amazon-photo can be acquired via the dataset interface of PyG \cite{fey2019a} \footnote{\url{https://pytorch-geometric.readthedocs.io/en/latest/modules/datasets.html}}. And ACM, DBLP, UAT, and Wiki are hosted by \citet{liu2023SurveyDeepGraph}\footnote{\url{https://github.com/yueliu1999/Awesome-Deep-Graph-Clustering}}.

\subsection{Baseline Implementation}
We adopt the open source implementations of baselines.
\begin{itemize}
	\item K-means: Faiss implementation\footnote{\url{https://github.com/facebookresearch/faiss}}  
	\item DEC: official implementation\footnote{\url{https://github.com/piiswrong/dec}}
	\item SDCN, DAEGC, and DFCN: the third party implementation\footnote{\url{https://github.com/Marigoldwu/A-Unified-Framework-for-Deep-Attribute-Graph-Clustering}}
	\item HSAN: official implementation\footnote{\url{https://github.com/yueliu1999/HSAN}}
	\item SCGC:  official implementation\footnote{\url{https://github.com/yueliu1999/SCGC}}
	\item GRACE: official implementation\footnote{\url{https://github.com/CRIPAC-DIG/GRACE}}
	\item S$^3$GC: official implementation\footnote{\url{https://drive.google.com/corp/drive/folders/18B_eWbdVhOURZhqwoBSsyryb4WsiYLQK}} 
	\item DinkNet: official implementation\footnote{\url{https://github.com/yueliu1999/Dink-Net}}
\end{itemize}
Regarding DinkNet, we use the official pretrained model weights for the Cora, Citeseer, and Amazon-Photo datasets. For the other datasets, we pretrain and fine-tune DinkNet from scratch, selecting the optimal combination from 100 sets of hyperparameter settings.

For the other baselines, we use the official hyperparameters on all datasets. For datasets not reported in the original papers, we tune the baseline methods over 80 sets of hyperparameters and report the best results.

\subsection{THESAURUS Implementation}
We implement THESAURUS using \texttt{PyTorch 2.1} \cite{paszke2017AutomaticDifferentiationPyTorch}, \texttt{PyG 2.5} \cite{fey2019a}, and \texttt{Faiss-GPU 1.8} \cite{johnson2019BillionscaleSimilaritySearch,douze2024FaissLibrary}, which are compiled with \texttt{CUDA 12.1}. To ensure reproducibility, all graph convolution operators utilize \texttt{SparseTensor} from \texttt{torch-sparse 0.6}\footnote{https://github.com/rusty1s/pytorch\_sparse} for sparse matrix multiplications. We implement the FGW-OT \cite{titouan2019optimal} from scratch based on \texttt{PyTorch}.

The weights of neural network \(f_{\omega} \circ f_{\theta}\) and prototypes \(\mathbf{S}\) are initialized with Kaiming uniform initializer \cite{he2015DelvingDeepRectifiers}. And THESAURUS is trained with Adam optimizer \cite{kingma2017AdamMethodStochastic}.

\subsection{THESAURUS Hyperparameters}
The hyperparameters of THESAURUS for all used datasets are listed in Table \ref{tab:hyperparameters}, including:
\begin{itemize}
	\item \(S\): the number of prototypes 
	\item \(\alpha \): the weight of graph structure cost in FGW-OT
	\item \(\tau\): the softmax temperature of prediction in Eq. \eqref{eq:P1}
	\item  \(pe\): the edge drop rate of data augmentation
	\item  \(px\): the attribute drop rate of data augmentation
	\item  \(T\): the training epochs
	\item lr: the learning rate
	\item wd: the weight decay of Adam optimizer.
\end{itemize}
Moreover, the momentum weight for the prototype graph $\beta_1$ is consistently set to 0.99 across all datasets, while the momentum weight for the prototype marginal  $\beta_2$ is set to 0.999.

\begin{table}[h]
	\centering
	\fontsize{8.5pt}{10pt}\selectfont 	
	\begin{tabular}{crrrrr|rrr}
		\toprule 
		Dataset 	& \(S\) &  \(\alpha\) & \(\tau\) & $pe$ & $px$ & $T$ & lr	& wd\\
		\midrule
		Cora  		& 18 &	0.70	& 0.60	& 0.4 	&	0.4	& 200 &	5e-4	&5e-5\\
		Citeseer	& 27 &	0.55 	& 0.80 	& 0.1 	&	0.3	& 200 & 5e-4	&5e-4\\
		Pubmed  	& 98 &	0.65 	& 0.15 	& 0.2	&	0.0	& 200 & 1e-3	&5e-3\\
		A-photo 	& 63 &	0.45	& 0.60	& 0.3	&	0.3	& 200 & 1e-3&0\\
		CoraFull	& 494&	0.25	& 0.40	& 0.4	& 	0.4	& 100 & 5e-4&1e-4	\\
		\midrule
		ACM			& 15 &	0.25	& 0.25	& 0.5	&	0.4	& 220 & 5e-3& 5e-4 \\
		DBLP		& 57 &	0.25	& 0.2	& 0.5	&	0.3	& 200 & 1e-3& 5e-3\\
		UAT			& 5	 &  0.95	& 0.65	& 0.2	&	0.5	& 220 & 5e-3& 0\\
		Wiki		& 240&	0.75	& 0.25	& 0.3	&	0.5	& 50  & 1e-2& 5e-5\\
		\bottomrule
	\end{tabular}
	\caption{The hyperparameters of THESAURUS}
	\label{tab:hyperparameters}
\end{table}

\section{More Experiment Details}
\subsection{Evaluation Metrics}

\subsubsection{Clustering Accuracy and Macro-F1 Score}
Clustering accuracy (ACC) and Macro-F1 score (F1) are not directly applicable as clusters do not inherently have labels that can be directly compared with the ground truth. Hence, it typically involves finding the best match between cluster labels and the ground truth labels using an optimal relabeling strategy. This strategy involves solving a linear sum assignment (LSA) problem \citep{lovasz1986MatchingTheoryNorthHolland,crouse2016Implementing2DRectangular} where each cluster label is mapped to a ground truth label to maximize the number of correct predictions.

The cost matrix $\mathbf{M}\in \mathbb{R}^{C\times C}$ is constructed such that each element 
$\mathbf{M}_{i,j}$ represents the \enquote{cost} of assigning cluster $i$
to true label $j$. In the context of maximizing clustering accuracy, this cost is  defined inversely as the number of data points in cluster $i$ that belong to the true label 
$j$. Therefore, solving the problem involves maximizing the total number of correct classifications. Formally, we have
\begin{equation}
	\max_\sigma\sum_{i=1}^k\mathbf{M}_{i,\sigma(i)},
\end{equation}
where $\sigma$ is a permutation of the set $\{1,2,\dots,C\}$ and ${C}$ is the number of classes. A fast sovler for LSA problems is \texttt{scipy.optimize.linear\_sum\_assignment}. With the solution $\sigma$, the cluster predictions of all nodes are first mapped to class predictions and then the normal accuracy and Macro-F1 computed like in classification tasks are taken as ACC and F1 for clustering.

\subsubsection{NMI and ARI}
NMI measures the amount of information that is shared between the predicted clustering and the ground truth clustering. It is a normalization of the Mutual Information (MI) to scale the results between 0 (no mutual information) and 1 (perfect correlation). ARI is an adjustment of the Rand Index that corrects for the chance grouping of elements, providing a more accurate measure of how well the clustering has performed. For their formal definitions, please refer to any textbooks, e.g., \cite{gan2020DataClusteringTheory},  discussing clustering algorithms.

\subsection{Environment}
All experiments were conducted on two computers. The first is an Ubuntu 22.04 server equipped with an RTX 4090 GPU (24GB), an Intel i7-12700 CPU, and 64GB of RAM. The second is an Ubuntu 20.04 server equipped with an RTX 4090 GPU (24GB), two Intel Xeon Gold 6240C CPUs, and 126GB of RAM.

Both computers have the same Conda virtual environment, with \texttt{PyTorch 2.1}, \texttt{PyG 2.5}, and \texttt{Faiss-GPU 1.8}, all built on \texttt{CUDA 12.1}.

\subsection{The Results on ACM, DBLP, UAT, and Wiki}
To follow the evaluation protocol of Dink-Net \cite{liu2023DinkNetNeuralClustering} and meet the page space limit, we only report the results on five datasets in the main content (Table \ref{tab:MainResults}). And to align with Dink-Net, we only report the results of one run in the main content. 

To assess the stability of THESAURUS, we carry out experiments on ACM, DBLP, UAT, and Wiki, computing the mean and standard deviation across five runs. The results, displayed in Tables \ref{tab:ExtraResutls1} and \ref{tab:ExtraResutls2}, demonstrate that THESAURUS consistently outperforms other methods across all four metrics. Moreover, the observed standard deviations are within a reasonable range, further confirming the stability of THESAURUS.

\begin{table*}[ht]
	\centering%
	\fontsize{9pt}{10pt}\selectfont 	
	\begin{tabular}{ccccccccc}
		\toprule 
		Datasets & Metrics & K-means & DEC & SDCN & DAEGC & DFCN & DCRN & Ours\tabularnewline
		\midrule
		\multirow{4}{*}{ACM} & ACC & 67.31\textpm 0.71 & 84.33\textpm 0.76 & 90.45\textpm 0.18 & 86.94\textpm 2.83 & 90.90\textpm 0.20 & 91.72\textpm 1.44 & \textbf{92.52\textpm 0.18}\tabularnewline
		& NMI & 32.44\textpm 0.46 & 54.54\textpm 1.51 & 68.31\textpm 0.25 & 56.18\textpm 4.15 & 69.40\textpm 0.40 & 71.41\textpm 1.27 & \textbf{73.26\textpm 0.27}\tabularnewline
		& ARI & 30.60\textpm 0.69 & 60.64\textpm 1.87 & 73.91\textpm 0.40 & 59.35\textpm 3.89 & 74.90\textpm 0.40 & 77.18\textpm 1.28 & \textbf{79.06\textpm 0.45}\tabularnewline
		& F1 & 67.57\textpm 0.74 & 84.51\textpm 0.74 & 90.42\textpm 0.19 & 87.07\textpm 2.79 & 90.80\textpm 0.20 & 83.92\textpm 0.76 & \textbf{92.54\textpm 0.18}\tabularnewline
		\midrule 
		\multirow{4}{*}{DBLP} & ACC & 38.65\textpm 0.65 & 58.16\textpm 0.56 & 68.05\textpm 1.81 & 62.05\textpm 0.48 & 75.00\textpm 0.31 & 60.86\textpm 1.53 & \textbf{82.25\textpm 0.32}\tabularnewline
		& NMI & 11.45\textpm 0.38 & 29.51\textpm 0.28 & 39.50\textpm 1.34 & 32.49\textpm 0.45 & 42.70\textpm 1.03 & 25.87\textpm 0.41 & \textbf{53.10\textpm 0.38}\tabularnewline
		& ARI & 6.97\textpm 0.39 & 23.92\textpm 0.39 & 39.15\textpm 2.01 & 21.03\textpm 0.52 & 45.98\textpm 1.20 & 22.05\textpm 0.73 & \textbf{59.29\textpm 0.62}\tabularnewline
		& F1 & 31.92\textpm 0.27 & 59.38\textpm 0.51 & 67.71\textpm 1.51 & 61.75\textpm 0.67 & 74.21\textpm 0.89 & 62.51\textpm 0.48 & \textbf{81.71\textpm 0.33}\tabularnewline
		\midrule 
		\multirow{4}{*}{UAT} & ACC & 42.47\textpm 0.15 & 45.61\textpm 1.84 & 52.25\textpm 1.91 & 52.29\textpm 0.49 & 33.61\textpm 0.09 & 49.92\textpm 1.25 & \textbf{62.25\textpm 0.92}\tabularnewline
		& NMI & 22.39\textpm 0.69 & 16.63\textpm 2.39 & 21.61\textpm 1.26 & 21.33\textpm 0.44 & 26.49\textpm 0.41 & 24.09\textpm 0.53 & \textbf{30.93\textpm 0.95}\tabularnewline
		& ARI & 15.71\textpm 0.76 & 13.14\textpm 1.97 & 21.63\textpm 1.49 & 20.50\textpm 0.51 & 11.87\textpm 0.23 & 17.17\textpm 0.69 & \textbf{31.12\textpm 1.33}\tabularnewline
		& F1 & 36.12\textpm 0.22 & 44.22\textpm 1.51 & 45.59\textpm 3.54 & 50.33\textpm 0.64 & 25.79\textpm 0.29 & 44.81\textpm 0.87 & \textbf{61.29\textpm 0.71}\tabularnewline
		\midrule 
		\multirow{4}{*}{Wiki} & ACC & 31.92\textpm 2.42 & 33.58\textpm 0.95 & 42.12\textpm 0.28 & 38.14\textpm 0.52 & 44.37\textpm 0.75 & 48.52\textpm 1.31 & \textbf{57.84\textpm 1.91}\tabularnewline
		& NMI & 29.4\textpm 3.20 & 30.51\textpm 0.82 & 40.95\textpm 0.62 & 31.42\textpm 0.33 & 41.94\textpm 1.13 & 45.81\textpm 0.27 & \textbf{53.41\textpm 0.96}\tabularnewline
		& ARI & 4.11\textpm 1.43 & 14.19\textpm 0.59 & 27.26\textpm 0.38 & 17.94\textpm 0.21 & 27.33\textpm 0.49 & 27.38\textpm 0.57 & \textbf{40.63\textpm 2.59}\tabularnewline
		& F1 & 20.29\textpm 1.32 & 22.38\textpm 0.47 & 37.55\textpm 0.54 & 24.51\textpm 0.19 & 37.53\textpm 0.72 & 37.47\textpm 0.63 & \textbf{47.84\textpm 1.86}\tabularnewline
		\bottomrule
	\end{tabular}
	
	\caption{The performance (\%) mean and standard deviation over 5 runs. The
		best is in bold.}
	\label{tab:ExtraResutls1}
\end{table*}

\begin{table*}[ht]
	\centering%
	\fontsize{9pt}{10pt}\selectfont 
	\begin{tabular}{cccccccc}
		\toprule 
		Datasets & Metrics & HSAN & S$^3$GC & SCGC & Dink-Net{*} & Dink-Net & Ours\tabularnewline
		\midrule
		\multirow{4}{*}{ACM} & ACC & 89.79\textpm 0.16 & 89.08\textpm 0.53 & 89.97\textpm 0.56 & 90.11\textpm 0.44 & 90.54\textpm 0.02 & \textbf{92.52\textpm 0.18}\tabularnewline
		& NMI & 66.97\textpm 0.50 & 64.22\textpm 1.10 & 67.08\textpm 0.97 & 68.32\textpm 0.93 & 69.00\textpm 0.05 & \textbf{73.26\textpm 0.27}\tabularnewline
		& ARI & 72.41\textpm 0.37 & 70.32\textpm 1.29 & 72.83\textpm 1.27 & 73.25\textpm 1.04 & 74.24\textpm 0.03 & \textbf{79.06\textpm 0.45}\tabularnewline
		& F1 & 89.73\textpm 0.18 & 89.08\textpm 0.53 & 89.89\textpm 0.58 & 90.06\textpm 0.43 & 90.48\textpm 0.02 & \textbf{92.54\textpm 0.18}\tabularnewline
		\midrule 
		\multirow{4}{*}{DBLP} & ACC & 73.18\textpm 0.77 & 76.89\textpm 0.31 & 66.82\textpm 1.45 & 73.31\textpm 2.41 & 74.71\textpm 0.02 & \textbf{82.25\textpm 0.32}\tabularnewline
		& NMI & 43.60\textpm 0.45 & 44.74\textpm 0.53 & 38.05\textpm 1.40 & 43.17\textpm 2.32 & 44.61\textpm 0.13 & \textbf{53.10\textpm 0.38}\tabularnewline
		& ARI & 44.80\textpm 0.80 & 49.14\textpm 0.67 & 35.42\textpm 1.82 & 43.51\textpm 3.26 & 45.55\textpm 0.86 & \textbf{59.29\textpm 0.62}\tabularnewline
		& F1 & 72.84\textpm 0.76 & 76.47\textpm 0.25 & 66.60\textpm 1.50 & 73.20\textpm 2.28 & 74.65\textpm 0.54 & \textbf{81.71\textpm 0.33}\tabularnewline
		\midrule 
		\multirow{4}{*}{UAT} & ACC & 56.04\textpm 0.67 & 49.55\textpm 4.36 & 56.58\textpm 1.62 & 54.84\textpm 1.83 & 58.07\textpm 0.17 & \textbf{62.25\textpm 0.92}\tabularnewline
		& NMI & 26.99\textpm 2.11 & 22.99\textpm 0.22 & 28.07\textpm 0.71 & 24.51\textpm 1.44 & 26.60\textpm 0.03 & \textbf{30.93\textpm 0.95}\tabularnewline
		& ARI & 25.22\textpm 1.96 & 20.49\textpm 1.36 & 24.80\textpm 1.85 & 23.45\textpm 1.55 & 25.66\textpm 0.06 & \textbf{31.12\textpm 1.33}\tabularnewline
		& F1 & 54.20\textpm 1.84 & 46.19\textpm 5.54 & 55.52\textpm 0.87 & 52.32\textpm 2.08 & 55.55\textpm 0.11 & \textbf{61.29\textpm 0.71}\tabularnewline
		\midrule 
		\multirow{4}{*}{Wiki} & ACC & 53.10\textpm 0.72 & 47.03\textpm 1.59 & 56.63\textpm 0.30 & 52.99\textpm 0.95 & 53.26\textpm 0.32 & \textbf{57.84\textpm 1.91}\tabularnewline
		& NMI & 50.11\textpm 0.32 & 22.49\textpm 0.59 & 52.62\textpm 0.49 & 49.91\textpm 1.11 & 51.06\textpm 0.27 & \textbf{53.41\textpm 0.96}\tabularnewline
		& ARI & 35.59\textpm 0.33 & 18.57\textpm 1.15 & 36.91\textpm 1.01 & 35.57\textpm 1.41 & 37.10\textpm 0.95 & \textbf{40.63\textpm 2.59}\tabularnewline
		& F1 & 44.83\textpm 0.71 & 44.54\textpm 3.94 & 47.52\textpm 0.89 & 44.77\textpm 0.44 & 45.01\textpm 0.39 & \textbf{47.84\textpm 1.86}\tabularnewline
		\bottomrule
	\end{tabular}
	
	\caption{The performance (\%) mean and standard deviation over 5 runs. The
		best is in bold.}
	\label{tab:ExtraResutls2}
\end{table*}

\subsection{More Visualization Results}
To elucidate the advantages of THESAURUS, we visualize the representations utilized for clustering along with the resultant cluster centers from both the previous SOTA, Dink-Net, and THESAURUS. This visualization is facilitated by dimensionality reduction to 2D using t-SNE, enabling detailed visual analysis.

\begin{figure*}[h]
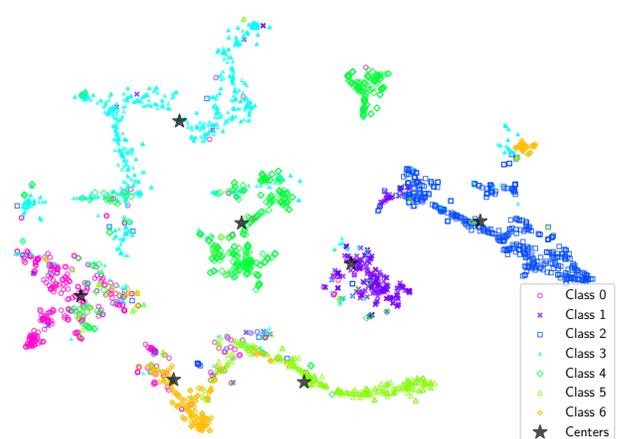

	\centering
	\begin{subfigure}{0.49\textwidth}
		\includegraphics[width=1\linewidth]{t-sne_cora_DinkNet}
		\caption{Dink-Net on Cora}
		\label{fig:LargeTSNECoraDinkNetCora}
	\end{subfigure}
	\begin{subfigure}{0.49\textwidth}
		\includegraphics[width=1\linewidth]{t-sne_cora_THESAURUS}
		\caption{THESAURUS on Cora}
		\label{fig:LargeTSNECoraTHESAURUSCora}
	\end{subfigure}
	\caption{The visualization of Dink-Net and THESAURUS on Cora, expanded from Fig. \ref{fig:Visual2D}.}
	\label{fig:LargeTSNECora}
\end{figure*}

\begin{figure*}[h]
	\centering
	\begin{subfigure}{0.49\textwidth}
		\includegraphics[width=1\linewidth]{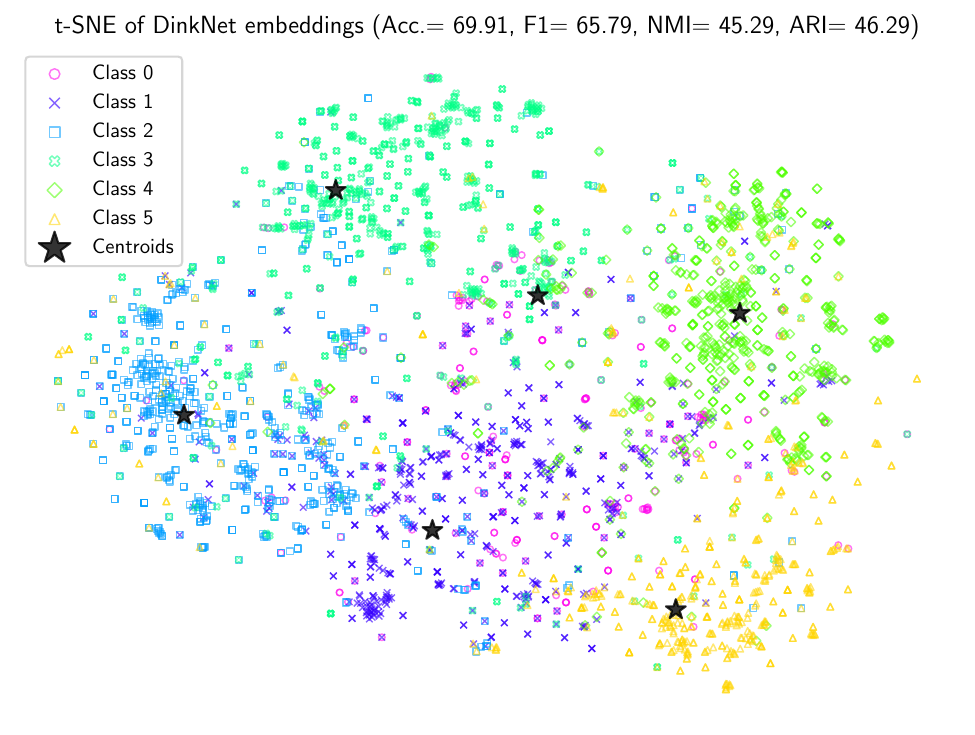}
		\caption{Dink-Net on Citeseer}
		\label{fig:LargeTSNECoraDinkNetCiteseer}
	\end{subfigure}
	\begin{subfigure}{0.49\textwidth}
		\includegraphics[width=1\linewidth]{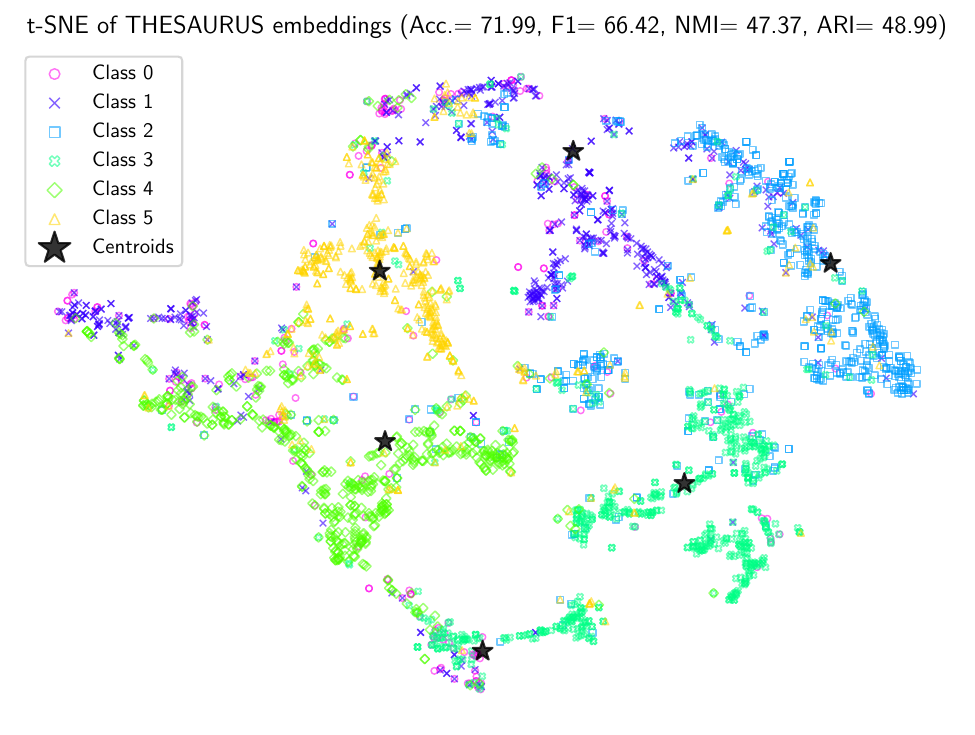}
		\caption{THESAURUS on Citeseer}
		\label{fig:LargeTSNECoraTHESAURUSCiteseer}
	\end{subfigure}
	
	\begin{subfigure}{0.49\textwidth}
		\includegraphics[width=1\linewidth]{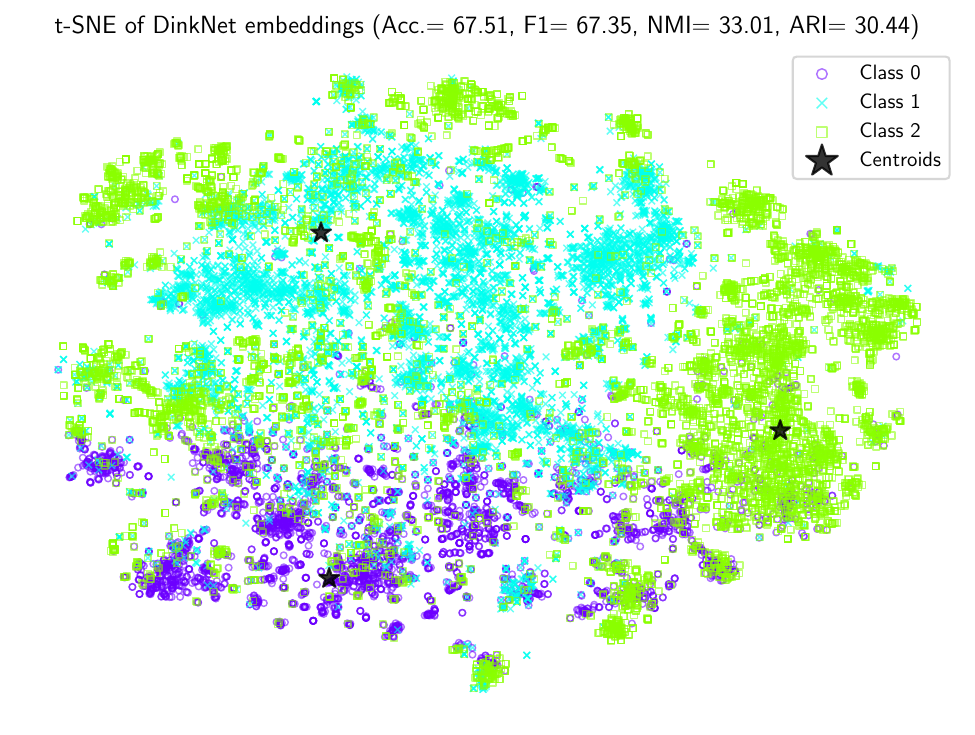}
		\caption{Dink-Net on Pubmed}
		\label{fig:LargeTSNECoraDinkNetPubmed}
	\end{subfigure}
	\begin{subfigure}{0.49\textwidth}
		\includegraphics[width=1\linewidth]{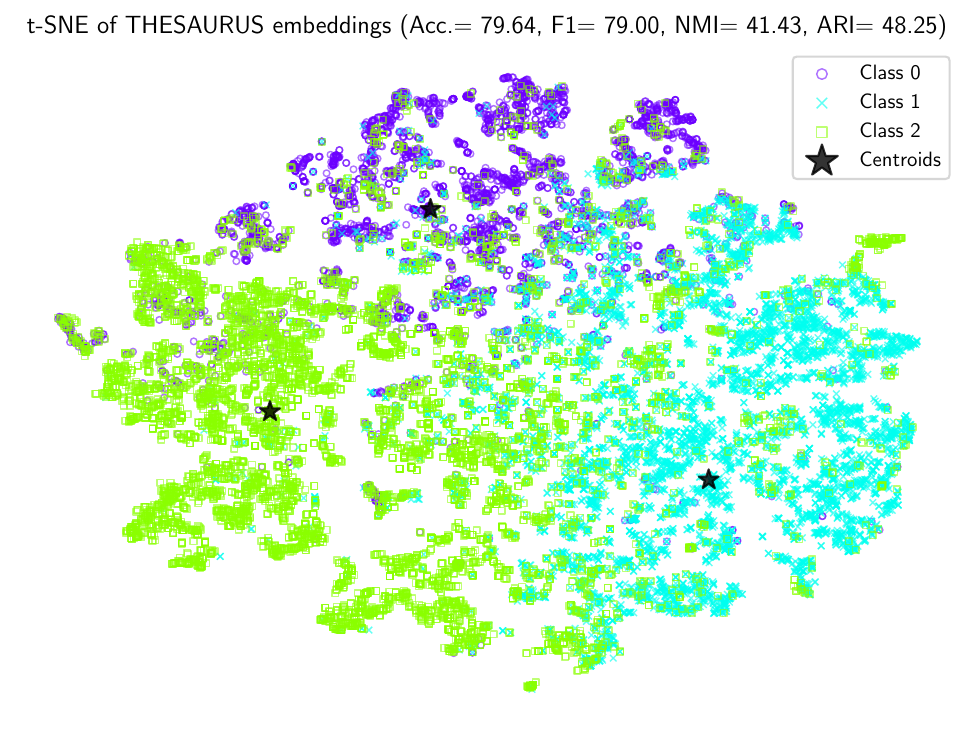}
		\caption{THESAURUS on Pubmed}
		\label{fig:LargeTSNECoraTHESAURUSPubmed}
	\end{subfigure}
	
	\begin{subfigure}{0.49\textwidth}
		\includegraphics[width=1\linewidth]{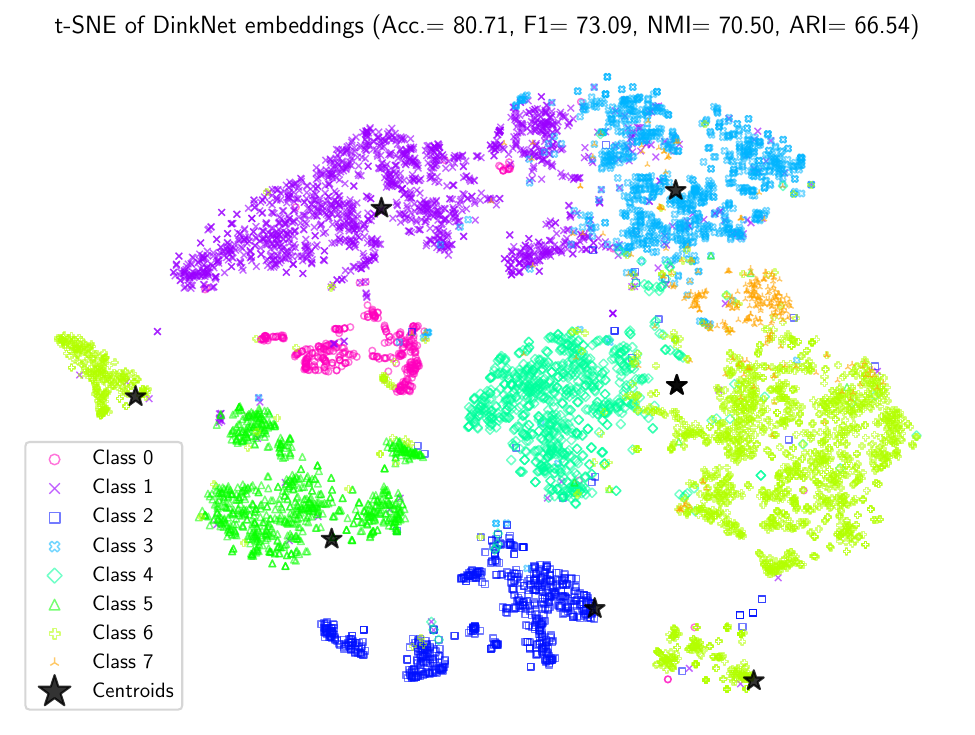}
		\caption{Dink-Net on Amazon-Photo}
		\label{fig:LargeTSNECoraDinkNetPhoto}
	\end{subfigure}
	\begin{subfigure}{0.49\textwidth}
		\includegraphics[width=1\linewidth]{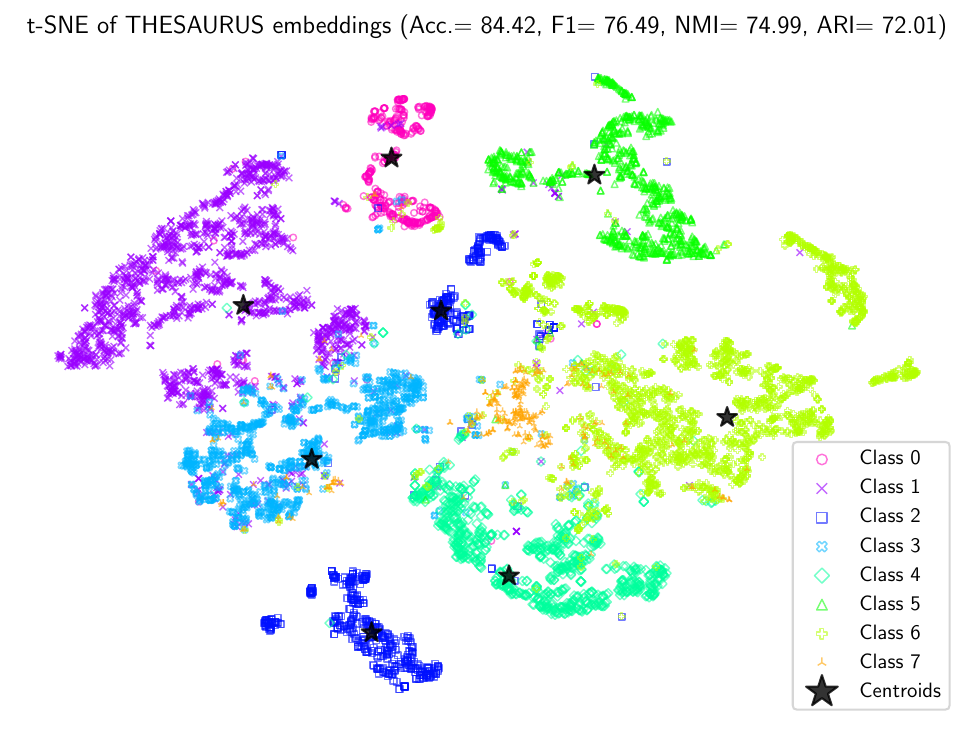}
		\caption{THESAURUS on  Amazon-Photo}
		\label{fig:LargeTSNECoraTHESAURUSPhoto}
	\end{subfigure}
	\caption{The visualization comparison between Dink-Net and THESAURUS on Citeseer, Pubmed, and Amazon-photo}
	\label{fig:LargeTSNE_group1}
\end{figure*}

\begin{figure*}[h]
	\centering
	
	\begin{subfigure}{0.49\textwidth}
		\includegraphics[width=1\linewidth]{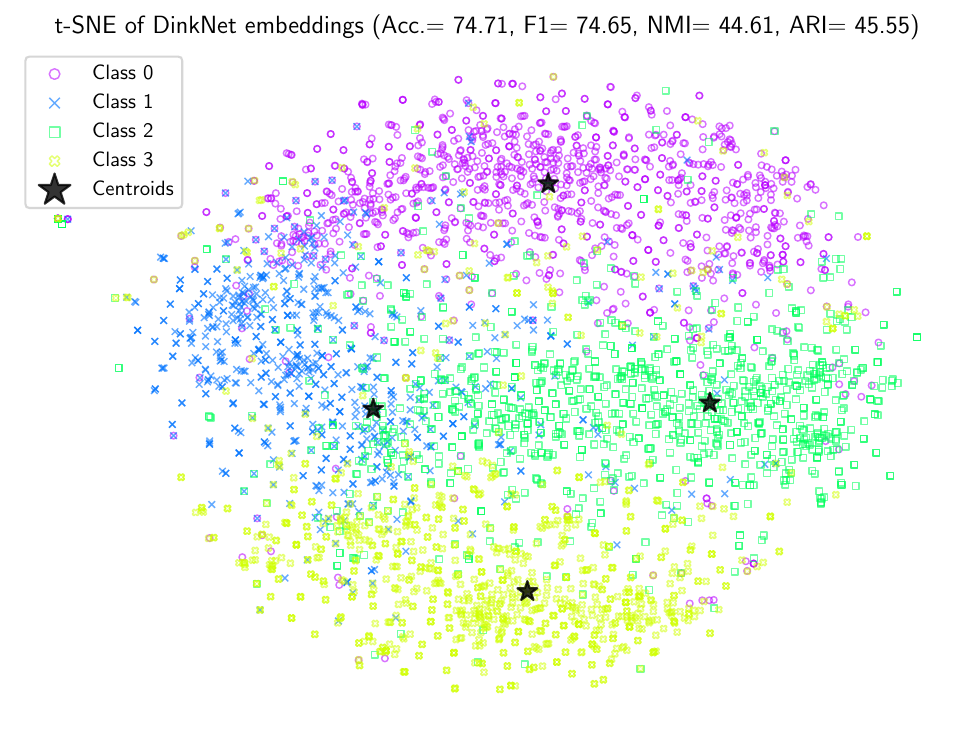}
		\caption{Dink-Net on DBLP}
		\label{fig:LargeTSNECoraDinkNetDBLP}
	\end{subfigure}
	\begin{subfigure}{0.49\textwidth}
		\includegraphics[width=1\linewidth]{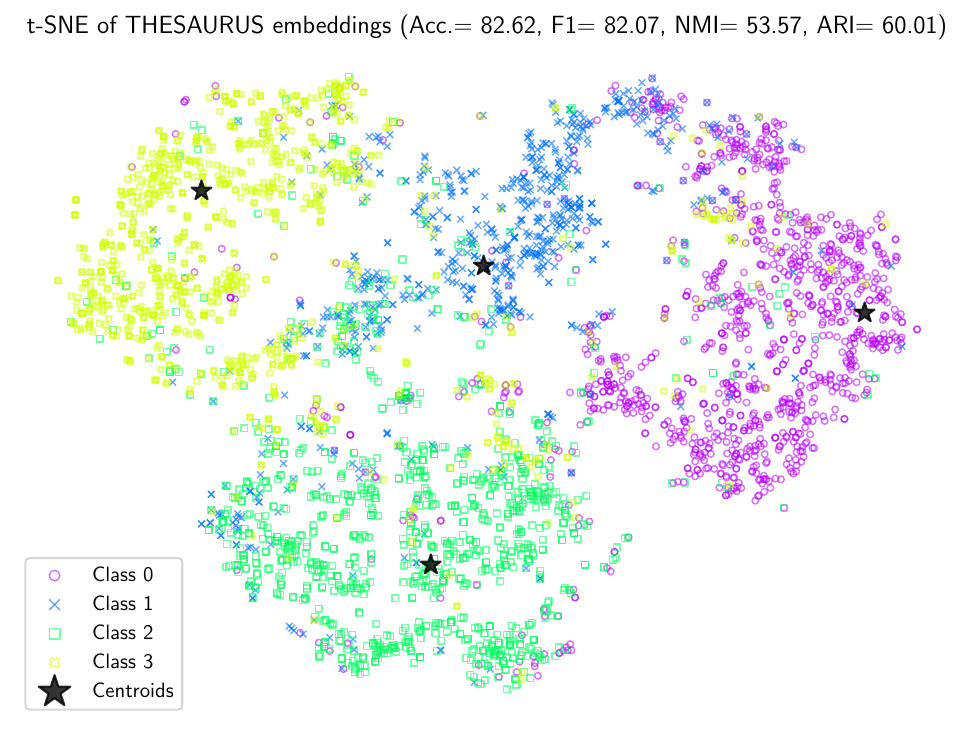}
		\caption{THESAURUS on DBLP}
		\label{fig:LargeTSNECoraTHESAURUSDBLP}
	\end{subfigure}
	
	\begin{subfigure}{0.49\textwidth}
		\includegraphics[width=1\linewidth]{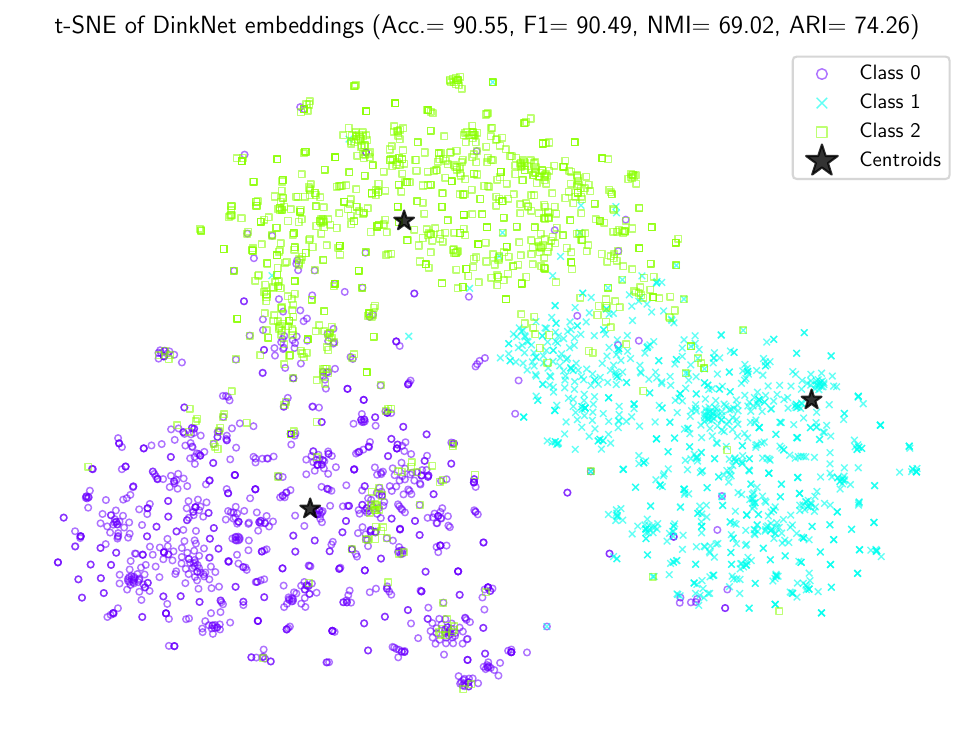}
		\caption{Dink-Net on ACM}
		\label{fig:LargeTSNECoraDinkNetACM}
	\end{subfigure}
	\begin{subfigure}{0.49\textwidth}
		\includegraphics[width=1\linewidth]{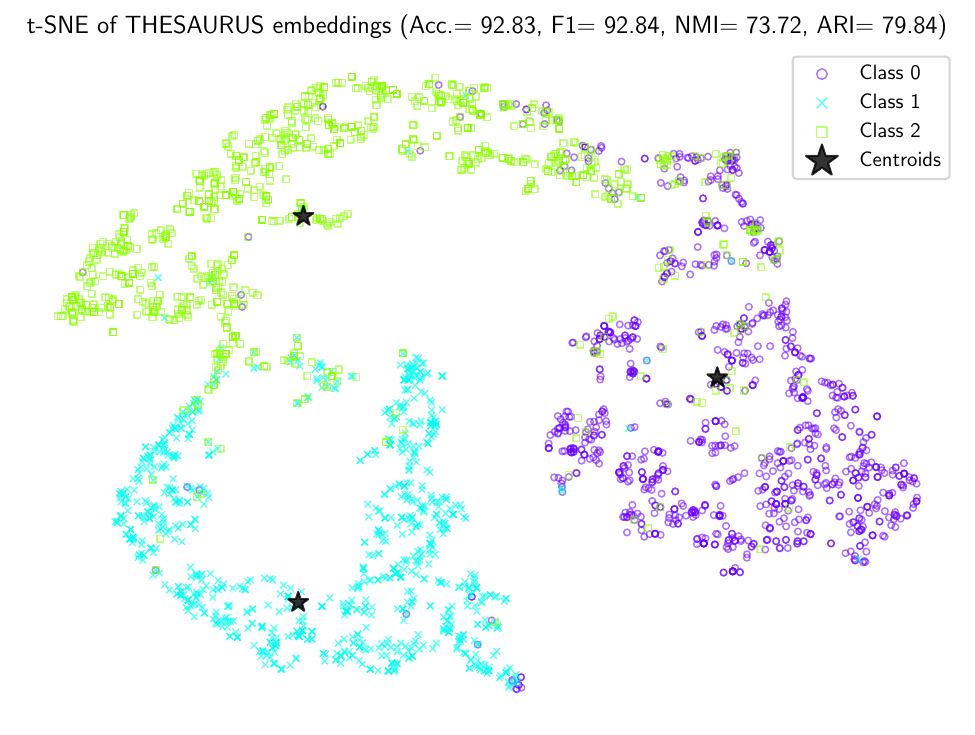}
		\caption{THESAURUS on ACM}
		\label{fig:LargeTSNECoraTHESAURUSACM}
	\end{subfigure}
	
	\begin{subfigure}{0.49\textwidth}
		\includegraphics[width=1\linewidth]{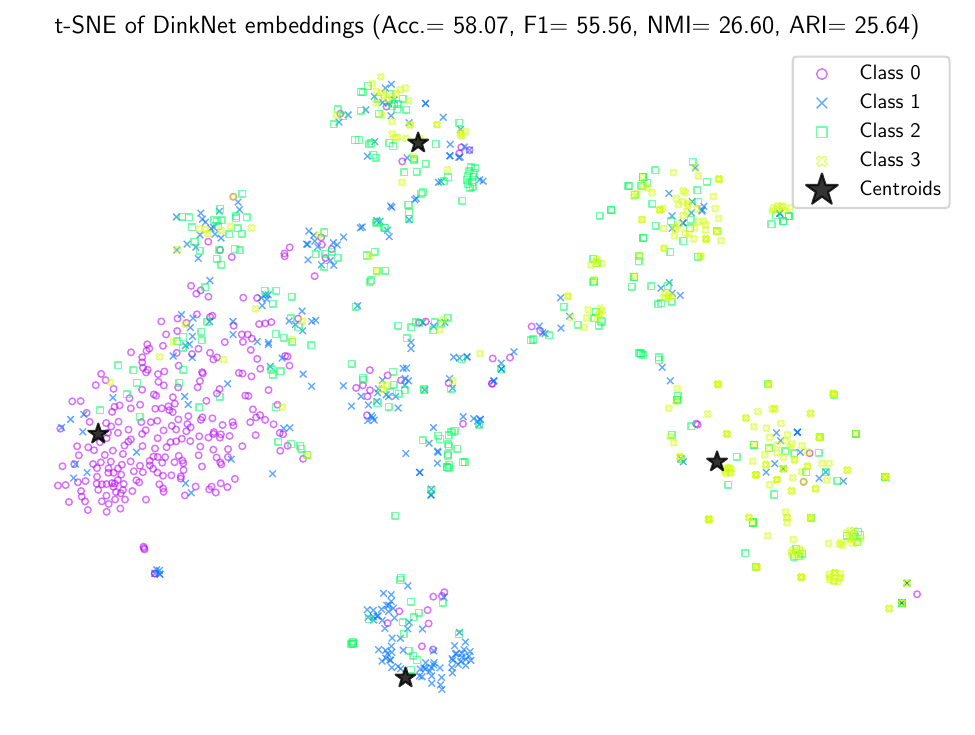}
		\caption{Dink-Net on UAT}
		\label{fig:LargeTSNECoraDinkNetUAT}
	\end{subfigure}
	\begin{subfigure}{0.49\textwidth}
		\includegraphics[width=1\linewidth]{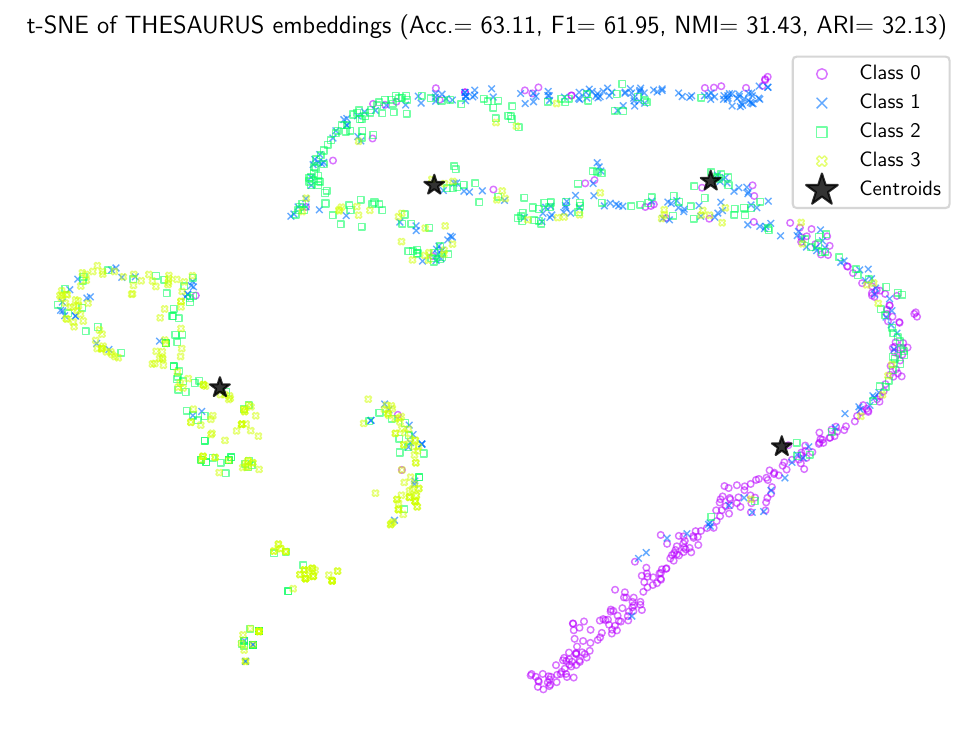}
		\caption{THESAURUS on UAT}
		\label{fig:LargeTSNECoraTHESAURUSUAT}
	\end{subfigure}
	\caption{The visualization of Dink-Net and THESAURUS on DBLP, ACM, and UAT. Only one run is shown.}
	\label{fig:LargeTSNE_group2}
\end{figure*}

A concise comparison between Dink-Net and THESAURUS on Cora is initially presented in Fig. \ref{fig:Visual2D}, resized to conform to the page constraints of the main text. An expanded version of this visualization is provided in Fig. \ref{fig:LargeTSNECora}, along with additional comparisons across various datasets presented in Fig. \ref{fig:LargeTSNE_group1} and Fig. \ref{fig:LargeTSNE_group2}. 
We do not visualize the results for CoraFull and Wiki due to the large number of classes, which makes clear representation visualization difficult.

All t-SNE reduction experiments are performed using \texttt{sklearn.manifold.TSNE(n\_components=2, learning\_rate="auto", random\_state=2050)} \cite{pedregosa2011ScikitlearnMachineLearning} and \texttt{Matplotlib} \cite{hunter2007Matplotlib2DGraphics}. These visualizations clearly illustrate that the representation space developed by THESAURUS offers superior cluster separability over Dink-Net, characterized by larger inter-cluster distances and smaller intra-cluster distances.

\end{document}